\documentclass[10pt,twocolumn,letterpaper]{article}

\usepackage[pagenumbers]{wacv}
\usepackage{stfloats}          
\definecolor{wacvblue}{rgb}{0.21,0.49,0.74}
\usepackage[accsupp]{axessibility}
\usepackage[pagebackref,breaklinks,colorlinks,allcolors=wacvblue]{hyperref}

\def\wacvPaperID{452}
\def\confName{WACV}
\def\confYear{2027}

\title{METRIC-GS: Selective Metric-Scale Estimation for 3D Gaussian Splatting}

\author{Sooyeun Yang$^{2}$ \qquad Jongseong Brad Choi$^{1,2,3}$\thanks{Corresponding author.}\\
$^{1}$Department of Mechanical Engineering, SUNY Korea\\
$^{2}$Department of Mechanical Engineering, Stony Brook University\\
$^{3}$Department of Computer Science, SUNY Korea\\
{\tt\small sooyeun.yang@stonybrook.edu, jongseong.choi@sunykorea.ac.kr}
}

\makeatletter
\apptocmd{\@maketitle}{%
  \begin{center}
    \includegraphics[width=0.80\linewidth]{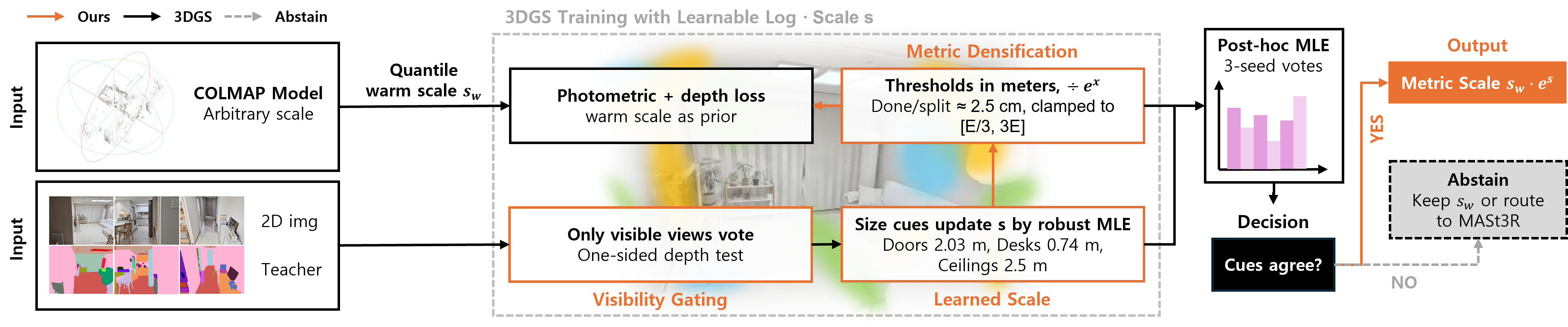}\par
    \vskip\abovecaptionskip
    \captionof{figure}{\textbf{Overview.} A quantile warm scale $s_w$ initializes the
  COLMAP model; 3DGS trains with a global log-scale $s$.
  A frozen 2D teacher labels Gaussians by \emph{visibility-aware}
  projection voting (one-sided depth test); instance-level size cues
  update $s$ by a robust MLE under architectural priors, and
  densification thresholds are meter-valued, converted through $e^{s}$.
  After training, a multi-seed post-hoc MLE reports the final metric
  scale $s_w e^{s}$, or \emph{abstains}, keeping $s_w$ or routing the
  scene to a heavyweight metric model.}
    \label{fig:pipeline}
  \end{center}
  \vspace{6pt}
}{}{\errmessage{could not patch maketitle}}
\makeatother

\begin{document}
\maketitle
\begin{abstract}
A 3D Gaussian Splatting (3DGS) reconstruction inherits arbitrary units
from structure-from-motion, so it cannot be measured or used to place AR
content until someone supplies a scale. Both standard recipes are
fragile on real captures: on five indoor scenes with tape-measured
ceilings a quantile floor-to-ceiling heuristic errs by 27\% and
metric-depth alignment by 35--71\% across three networks. We present
METRIC-GS, a
\emph{selective} scale estimator for real photographs of indoor scenes:
it reads the scale off the reconstruction's own semantic content and,
just as importantly, refuses when the evidence is thin. A frozen 2D
teacher labels Gaussians by visibility-aware voting; instance-level
size priors for architectural and furniture classes (doors, tables, the
floor-to-ceiling gap) are fused by a robust maximum-likelihood step;
and an inlier-mass test turns thin evidence into an explicit abstention
that doubles as a router to a heavyweight metric model. On 47 ScanNet++
scenes with laser ground truth, under a configuration frozen before any
public evaluation, the estimator alone cuts the heuristic's mean error from
21.1\% to 16.6\% (9.3\% on the 28 scenes it acts on) at ${\sim}11$\,s
per scene. MASt3R remains the more accurate standalone estimator
(9.9\%), but routing only our refusals to it matches running it
everywhere with 60\% fewer invocations; this selective hybrid is the
configuration we advocate. Visibility-aware voting is what makes the
estimator actionable, raising the scenes it can act on from 16 to 28
without changing deployed accuracy. Applied unchanged to 13 phone and
drone captures of our own, on which MASt3R fails its own registration
gate 12 times, the system improves all five captures it acts on and
degrades none. We also report that a current feed-forward metric model scores
2.3\% on a benchmark inside its training mixture and 28--35\% on
ordinary captures it has never seen: a published metric-scale number is
a best case, not a deployment estimate.
\end{abstract}

\section{Introduction}
\label{sec:intro}

3D Gaussian Splatting (3DGS)~\cite{kerbl3Dgaussians} reconstructs an
interior beautifully, but in whatever units structure-from-motion
happened to produce. For novel-view synthesis the ambiguity is harmless.
The moment the reconstruction has to support a measurement, a robot's
path, or an AR object placed at the right size, someone has to supply a
scale --- and today that someone is a human with a tape measure. A phone
that keeps its inertial stream answers this for free, so the problem
belongs to the captures that cannot: DSLR sessions, photo collections,
drone surveys, and any pipeline reconstructing from images alone.

\begin{figure}[t]
  \centering
  \includegraphics[width=0.52\linewidth]{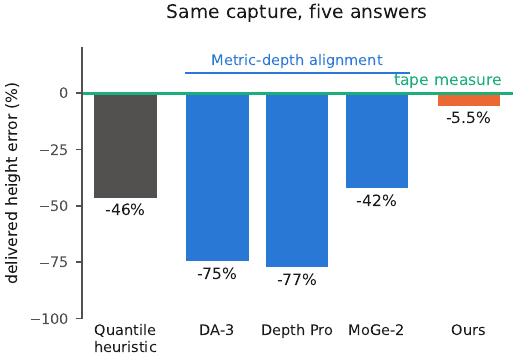}
  \caption{\textbf{Scale recovery is not solved.} Delivered ceiling
  height on one tape-measured capture (\texttt{room2}; all five in the
  supplement): the quantile heuristic and metric-depth alignment disagree
  with the tape measure and, across three networks, with each other by a
  factor of $2.5$. Our robust fusion of semantic size cues is within
  $6\%$.}
  \label{fig:teaser}
\end{figure}

Two off-the-shelf recipes exist. A geometric heuristic scales the
reconstruction so that its floor-to-ceiling extent matches a known room
height; a learned alternative aligns SfM depth to a monocular metric-depth
network. Both are fragile on real captures: on five indoor scenes with
tape-measured ceiling heights, the quantile heuristic errs by 27\% on
average and metric-depth alignment by 35--71\% \emph{across three networks}, a systematic indoor
depth bias rather than one model's flaw; the three networks also
disagree with \emph{each other} by up to a factor of $2.5$
(Fig.~\ref{fig:teaser}).

Our key observation is that an indoor reconstruction contains many
\emph{semantic metric anchors} (doors, furniture, the floor-to-ceiling
gap) whose real-world sizes are standardized or tightly distributed;
measured per instance on semantically labeled Gaussians, they
over-constrain the global scale. This fixes the operating regime, and we
state it upfront: METRIC-GS targets \emph{real photographs of indoor
scenes whose architecture is visible}, with hand-specified size priors
drawn from building standards and a frozen 2D segmenter as the only
learned component. Outside that regime (outdoor, synthetic appearance,
close-ups that never frame a wall) there is nothing to fuse, and the
method's contract is to say so rather than to guess.

Concretely, a frozen 2D teacher labels Gaussians by projection voting,
instance clusters are measured, and a robust maximum-likelihood step
fuses the measurements under the size priors (Sec.~\ref{sec:method}).
Two design decisions carry the system. First, reliable labeling requires the
reconstruction itself: naive projection voting degrades as views are
added, because occluded points inherit their occluders' labels. Our
\emph{visibility-aware voting} filters each vote by the model's own
rendered depth. Its effect is one of \emph{coverage}, not accuracy:
over all 47 test scenes it lifts the number the estimator can act on
from 16 to 28, while the deployed mean error is statistically unchanged
(Sec.~\ref{sec:exp-e5}). Second, an \emph{inlier-mass test} refuses the
correction when too few cues agree, so thin evidence becomes an explicit
abstention rather than a confident wrong answer; the same signal routes
the scene to a heavyweight metric model.

On 47 ScanNet++ scenes with laser-derived metric ground truth, under a
configuration frozen before public evaluation, the estimator alone improves
the quantile heuristic's mean error from $21.1\%$ to $16.6\%$ (median
$17.5\%\!\to\!10.6\%$; Wilcoxon $p{=}2.6{\times}10^{-3}$), reaching
$9.3\%$ on the 28 scenes it acts on at ${\sim}11$\,s overhead.
We are explicit about where the result is strongest: MASt3R is more
accurate as a standalone estimator ($9.9\%$) at over $20\times$ the
estimation cost, but on scenes where we act the two are statistically
indistinguishable, and routing only our 19 refusals to MASt3R matches
running it everywhere with $60\%$ fewer invocations.
That \emph{selective hybrid}, not the standalone accuracy, is the
contribution we advocate deploying. Applied unchanged to 13 captures we
recorded with consumer phones and a drone, the frozen system improves
every capture it acts on, abstains on the rest, and degrades none,
while MASt3R fails its own registration gate on 12 of the 13.

Our contributions are:
\begin{itemize}
  \item a robust fusion of instance-level semantic size cues that reads
        a metric reference off an indoor reconstruction itself; its
        measured contribution is robustness, removing the catastrophic
        failures single-cue estimators cannot defend against;
  \item \emph{visibility-aware voting}, which restricts votes to views
        that actually observe each Gaussian; it raises the scenes the
        estimator can act on from 16 to 28 at unchanged deployed
        accuracy, i.e.\ a coverage gain;
  \item \emph{abstention as a first-class output}: an inlier-mass test
        that converts insufficient evidence into an explicit refusal
        and doubles as a router, matching a heavyweight metric model's
        accuracy at 60\% fewer invocations;
  \item a measurement worth reporting independently of our estimator: a
        current feed-forward metric model scores $2.3\%$ on a benchmark
        inside its training mixture, $5.6\%$ on one it was evaluated
        against, and $28$--$35\%$ on captures it has never seen;
  \item a deployment study under development/test discipline: a
        configuration frozen before public evaluation, pre-registered
        held-out predictions, significance tests and negative results,
        on ScanNet++ (47 scenes), four further public benchmarks, and
        13 captures of our own.
\end{itemize}

\section{Related Work}
\label{sec:related}

\paragraph{3D Gaussian Splatting.}
3DGS~\cite{kerbl3Dgaussians} represents a scene as anisotropic Gaussians
optimized with photometric losses and grown by densification heuristics;
numerous variants refine its sampling and densification
behavior~\cite{mipsplatting, taming3dgs}. These heuristics (gradient
thresholds, world-size pruning, learning-rate schedules) are expressed in
world units, so they silently depend on the arbitrary scale that
structure-from-motion~\cite{schoenberger2016sfm} assigns to the scene. We
make this dependence explicit and turn it into a feature: densification
thresholds defined in meters couple scale accuracy to reconstruction
quality (Sec.~\ref{sec:method-loop}).

\paragraph{Recovering metric scale.}
Monocular metric depth estimators~\cite{depthanythingv2, metric3d, zoedepth}
enable a standard alignment recipe: fit a scale (and shift) between
predicted depth and SfM depth~\cite{vipe, depthanything3}. We show this
recipe carries the depth network's systematic indoor bias
(Sec.~\ref{sec:exp-dev}). Feed-forward reconstruction models trained with
metric supervision, such as DUSt3R~\cite{dust3r}, MASt3R~\cite{mast3r},
VGGT~\cite{vggt}, and more recently MapAnything~\cite{mapanything} and
AMB3R~\cite{amb3r}, recover impressive metric geometry from images
alone, but replace rather than augment an existing SfM/3DGS pipeline, and
provide no self-assessment of their scale reliability. We compare
empirically to MASt3R's metric checkpoint; MapAnything and AMB3R, its
direct successors, are not evaluated here, so the MASt3R error we report
is an upper bound on that family's error on curated benchmarks; its
successors may do better. Others in this family are explicitly scale-free:
$\pi^3$~\cite{pi3} predicts scale-invariant point maps, so a metric
anchor of the kind we supply remains necessary downstream.
MoGe-2~\cite{moge2} shares our factorization into relative geometry plus
a separate global scale, but predicts that scalar from a learned image
prior, whereas we read it off the reconstruction's own semantic content,
which is what makes an evidence-based refusal possible.
Monocular systems have long recovered scale from a geometric anchor such
as known object sizes or a ground plane at calibrated camera
height~\cite{scaleSLAM, objectSLAM, vgpnav}; a single anchor is precise
where it applies but fails when unobserved, whereas we fuse many
redundant anchors. Aligning monocular depth priors instead, as HI-SLAM2
does~\cite{hislam2}, imports the depth network's metric bias, which our
scale-invariant use of depth avoids.

\paragraph{Semantics in 3DGS.}
Distilling 2D semantics into Gaussians supports open-vocabulary querying
and editing~\cite{feature3dgs, langsplat}; structural priors
(planes, layout) regularize indoor geometry~\cite{planargs, gsroom, spcgs}.
These works target geometry quality or semantic access; none address the
metric scale. Closest to our voting are LabelGS~\cite{labelgs}, which
lifts 2D masks with explicit occlusion analysis so occluded Gaussians do
not inherit foreground labels, and VALA~\cite{vala}, which gates each ray
by a Gaussian's marginal contribution. Both address the same pathology,
but solve it for segmentation quality. Ours is a cheaper
one-sided test against rendered depth, aimed at unbiased instance
\emph{extents}; their finer visibility reasoning is a natural front-end
for our partial-instance failure mode.

\section{Method}
\label{sec:method}

\subsection{Problem, notation and overview}
\label{sec:method-overview}
3DGS optimizes a set of anisotropic Gaussians
$\Theta=\{(\boldsymbol{\mu}_i,\Sigma_i,o_i,c_i)\}$ against photometric
losses, growing and pruning the set with heuristics whose thresholds are
expressed in world units~\cite{kerbl3Dgaussians}. The COLMAP
reconstruction it consumes is defined only up to a global scale; our
task is to recover that scalar $\hat{s}$ (meters per model unit), or to
state explicitly that the evidence does not support one.

We use the following terms throughout. The \emph{warm estimate} $s_w$ is
a cheap geometric guess applied once, before training
(Sec.~\ref{sec:method-warm}); the \emph{residual} log-scale $s$ is the
correction we estimate, so the delivered scale is $\hat{s}=s_w e^{s}$.
A \emph{cue} is one measured size paired with a size prior; the
\emph{inlier mass} $M$ is the weight of cues that agree with the fitted
scale. A scene is \emph{active} if $M$ passes the gate and the estimator
reports its own $\hat{s}$, and \emph{abstained} if it refuses and
returns $s_w$ unchanged; the \emph{router} is the deployment that sends
exactly the abstained scenes to a heavyweight metric model.
\emph{Deployed error} is the scale error of whatever the system returns,
active or not. Figure~\ref{fig:pipeline} overviews the pipeline: after
the warm start, 3DGS trains while a frozen 2D teacher periodically
labels the Gaussians (Sec.~\ref{sec:method-voting}), instance sizes are
measured (Sec.~\ref{sec:method-cues}) and fused into $s$
(Sec.~\ref{sec:method-fusion}), which keeps densification in meters
(Sec.~\ref{sec:method-loop}); after training, a post-hoc fit of the
same objective reports $\hat{s}$ or abstains
(Sec.~\ref{sec:method-abstain}).

\subsection{Warm start}
\label{sec:method-warm}
The warm scale maps the reconstruction's dominant vertical extent, the
robust median over multi-quantile floor/ceiling spans of the point
heights, to a nominal room height ($2.5$\,m). This heuristic is fragile
(outlier points corrupt the span; Sec.~\ref{sec:exp-snpp}), so we keep
it only as the initialization and quadratic prior of the residual $s$.

\subsection{Visibility-aware semantic voting}
\label{sec:method-voting}
A frozen 2D teacher (Mask2Former R50 on ADE20K~\cite{mask2former}) labels
training views; projecting Gaussian centers into $V$ views and letting
each view vote yields per-Gaussian labels. Naively, \emph{more views make
this worse}: an occluded Gaussian projects into views that do not see it
and inherits its occluder's label, so the winner's vote share drops below
the ambiguity threshold as $V$ grows (in our diagnostics, usable cues fell
from 22 to 8 when doubling $V$). We therefore let a view vote only if the
model itself confirms visibility: with rendered depth $D_v$ at the
projection $(u_i^v, v_i^v)$ and camera depth $z_i^v$, the vote requires
\begin{equation}
  z_i^v \;<\; D_v(u_i^v,v_i^v)\,(1+\tau_{\mathrm{vis}}),
  \qquad \tau_{\mathrm{vis}}=0.05 .
  \label{eq:visibility}
\end{equation}
The test is deliberately \emph{one-sided}: a symmetric $|z-D|$ test also
rejects points at grazing incidence, i.e.\ floors, where per-pixel depth
blending shifts $D_v$ off the surface, and silently starves the floor
cues (15{,}856$\to$146 votes on one scene). A Gaussian keeps the winning
class if it holds at least half of its visibility-filtered votes. The
schedule is fixed: every 200 training iterations, 24 randomly drawn
views vote; the post-hoc estimate uses 48 views under each of three
fixed seeds. If fewer than 5\% of Gaussians end up labeled, the
tolerance is tripled once and then the test is dropped, so a degenerate
depth render cannot starve the estimator. Both design choices are
ablated over all 47 test scenes in Sec.~\ref{sec:exp-e5}; the gain is in
coverage, not accuracy.

\subsection{Instance-level metric size cues}
\label{sec:method-cues}
Classes with standardized or tightly distributed physical sizes (doors,
tables, toilets, the floor-to-ceiling gap) act as metric anchors. A class
is measured only if at least 50 Gaussians carry its label. Its Gaussians
are split into spatial instances by voxel-grid connected components
(26-neighborhood, voxel size $4\times$ the class' median
nearest-neighbor distance), keeping at most the four largest and
discarding fragments below half the largest instance's extent; each
instance yields a $[0.02,0.98]$ quantile-trimmed extent along the
vertical axis. Measurement biases are structural and we exploit their
opposing signs: a \emph{cluster} extent (the whole labeled set) is biased
up by label spillover, an \emph{instance} extent is biased down by
partial reconstruction, so the hybrid set lets the robust estimator
settle between them. Cluster cues are kept only for classes with at
most one instance (the union of several measures the room, not the
object), and instance and cluster cues share the class weight equally.
The floor$\to$ceiling gap is measured between histogram \emph{modes} of
the two height distributions, since contamination lies strictly on one
side of a planar surface, and is rejected if it falls below 15\% of the
labeled height span. The vertical axis itself is semantic: among the
floor-plane normal, ceiling-plane normal, and global-PCA candidates we
keep the one maximizing the floor--ceiling separation, since any tilt
attenuates it by the cosine; tilts up to $5^\circ$ cost nothing
measurable and larger ones drive the estimator into abstention
(supplement). The structural limitation of this cue model, and the principal
residual failure, is that a partially reconstructed or partially labeled
instance measures short, and nothing in the measurement itself
distinguishes that from a scale error (Sec.~\ref{sec:exp-snpp}).

\subsection{Robust fusion}
\label{sec:method-fusion}
Each cue $c$ contributes a measurement $m_c$ (model units) with prior
$\mathcal{N}(\mu_c,\sigma_c^2)$ (meters) and weight $w_c$ set once from
public architectural standards and never tuned per dataset. The 19 anchors
fall into three reliability tiers: code-regulated dimensions carry $w{=}1$
(door $2.03{\pm}0.10$\,m, floor$\to$ceiling $2.50{\pm}0.30$\,m),
ergonomically constrained furniture $w{\in}[0.4,0.6]$ (table
$0.74{\pm}0.08$, counter $0.90{\pm}0.10$), and style-dependent objects
$w{\le}0.3$ (wardrobe $2.00{\pm}0.30$, television $0.70{\pm}0.30$); the
full catalog is in the supplement. Since $\sigma_c$ is the spread of the
published range, a frequent class can still be nearly uninformative.
Cues are fused as if independent, which is conservative: correlated
under-measurement is precisely the failure mode above. With
$\rho$ the Geman--McClure penalty ($c{=}3$), the scale objective is
\begin{equation}
  E(s) \;=\; \sum_c w_c\,
  \rho\!\Big(\tfrac{e^{s} m_c - \mu_c}{\sigma_c}\Big)
  \;+\; \lambda \Big(\tfrac{s}{\sigma_w}\Big)^{2},
  \label{eq:fusion}
\end{equation}
where the second term is the warm prior on the residual
($\sigma_w{=}0.35$ in log-space, $\lambda{=}1$). It is deliberately
\emph{quadratic}: a robust prior saturates for far-away modes and stops
resisting when protection is needed.

\subsection{Scale in the training loop}
\label{sec:method-loop}
The complete training problem is the standard 3DGS objective plus a
scale-invariant depth term,
\begin{equation}
  \mathcal{L}(\Theta) = \mathcal{L}_{\mathrm{photo}}(\Theta)
  + \lambda_{\mathrm{d}}\,\mathcal{L}_{\mathrm{depth}}(\Theta),
  \qquad \lambda_{\mathrm{d}}=0.05,
  \label{eq:total}
\end{equation}
where $\mathcal{L}_{\mathrm{photo}}$ is the unmodified
$\mathcal{L}_1$\,+\,D-SSIM loss of~\cite{kerbl3Dgaussians}. $\mathcal{L}_{\mathrm{depth}}$ is a Huber
loss in log-space between rendered depth and cached monocular depth with
each image's median removed, so it shapes local geometry without letting
the depth network's metric bias touch the scale. $\mathcal{L}$ is
invariant to $s$ (a global rescale of scene and cameras leaves every
render unchanged), so the two unknowns are optimized in alternation:
$\Theta$ descends $\mathcal{L}$ at every iteration, and every 200
iterations, after a vote, $s$ is re-fitted by 25 Adam steps on
Eq.~\eqref{eq:fusion} over measurements buffered from the last $K{=}5$
votes. The coupling runs the other way: densification is anchored in
meters. We fix the scene extent $E_m = E\,e^{s_0}$ at initialization and
pass $E_m/e^{s}$, clamped to $[E/3,\,3E]$ so a runaway scale cannot
destroy geometry, wherever 3DGS uses the extent, so the clone/split
threshold is $1\%$ and the prune threshold $10\%$ of a \emph{metric}
extent ($0.95$--$7.4$\,m, median $2.5$\,m, on ScanNet++, i.e.\ a
$2.5$\,cm median split threshold).

This is also why the estimator is coupled to 3DGS rather than run on
the SfM point cloud. The cue measurement is a post-hoc operation on any
labeled point set, and the in-loop scale adds no accuracy
(Sec.~\ref{sec:exp-e5}). What 3DGS supplies is (i) a dense surface
whose \emph{rendered depth} makes the visibility test of
Eq.~\eqref{eq:visibility} possible, which a sparse SfM cloud cannot,
and (ii) meter-valued densification, which makes the delivered model
metric rather than a model plus a number.

\subsection{Post-hoc estimate and abstention}
\label{sec:method-abstain}
After training, the reported scale minimizes Eq.~\eqref{eq:fusion} on
cues pooled over three vote seeds, scanned exhaustively at 2001
log-spaced points over $e^{s}\in[0.05,20]$: a scan rather than a descent,
since $E$ is non-convex under a robust loss; it is deterministic and costs
under $10$\,ms. A correction is only trustworthy if enough cues agree, so
we measure the \emph{inlier mass} $M=\sum_{c:|r_c|<2} w_c$ at the
minimizer, with $r_c$ the normalized residual of Eq.~\eqref{eq:fusion},
and refuse the semantic correction when $M<M_{\min}{=}1.5$, returning
$s{=}0$ (the warm estimate). The threshold sits on a wide plateau and
needs no per-teacher recalibration (Sec.~\ref{sec:exp-e5}); across
capture regimes the decision tracks the teacher's 2D anchor-class
coverage, within a regime it reflects cue disagreement
(Sec.~\ref{sec:exp-regime}); and it is validated pre-registered on
held-out scenes (Sec.~\ref{sec:exp-heldout}).
Abstention also acts as the \emph{router}: scenes it flags are delegated
to a heavyweight metric reconstruction model (Tab.~\ref{tab:snpp}).

\section{Experiments}
\label{sec:exp}

\subsection{Setup}
\label{sec:exp-setup}
\paragraph{Development/test discipline.}
The method was designed and configured on five self-captured indoor scenes
(phone room-scans with tape-measured ceiling heights); every threshold in
Sec.~\ref{sec:method} was frozen there, before any public evaluation.
The first 19 ScanNet++ scenes were then evaluated once and serve as the
ablation split (Sec.~\ref{sec:exp-e5}, supplement); the remaining 28
and every held-out study ran afterwards, unchanged.

\paragraph{Protocols.}
We use three ground-truth protocols. \emph{(P1) Controlled ambiguity:} a
random sim(3) (log-uniform scale in $[0.25,4]$ and a random rotation,
seeded by a hash of the scene name) is applied to a laser-registered
metric model (ScanNet++~\cite{scannetpp}, ETH3D~\cite{eth3d},
Replica~\cite{replica}); the applied scale is the exact ground truth, and
the rotation removes gravity alignment so the semantic vertical axis is
genuinely exercised. \emph{(P2) Real SfM:} COLMAP~\cite{schoenberger2016sfm}
is run from the images; ground truth is the Umeyama-alignment scale of its
camera centers against the metric trajectory, gated at 20\,cm RMSE
(TUM~\cite{tum}, ARKitScenes~\cite{arkitscenes}, ScanNet++). \emph{(P3)
Delivered height:} on the development captures, the semantic
floor-to-ceiling gap of every method's final model against the tape
measure, with one shared measurement operator.

\paragraph{Baselines.}
The quantile floor/ceiling heuristic (our warm start, run standalone);
metric-depth alignment (median ratio between monocular metric depth and
SfM depth), run with \emph{three} networks, Depth Anything
3~\cite{depthanything3}, Depth Pro~\cite{depthpro} and MoGe-2~\cite{moge2},
so the reported failure reflects the recipe's indoor calibration bias,
common to all three (two also appear in Tab.~\ref{tab:snpp}; other
metric-depth models exist~\cite{metric3dv2,unidepth}); and
MASt3R~\cite{mast3r} with
its metric checkpoint, reconstructing $k{=}12$ evenly sampled frames by
sparse global alignment and reading the scale off a Umeyama fit to ground
truth ($|s{-}1|$). VGGT~\cite{vggt} predicts in a scale-normalized frame
and is therefore not a metric baseline. All timings are on one RTX~4080;
training runs 30k iterations ($\sim$6--16 min per scene), the post-hoc
estimator takes $\sim$11\,s (teacher inference, visibility renders,
and the seed ensemble all included; full breakdown in the supplement),
and the scale machinery adds no measurable
training overhead (one scene: 5.6 vs.\ 6.2\,min baseline). Nor does
it cost rendering quality: on the 19-scene development split the full
system reaches 32.62\,dB PSNR, 0.951 SSIM and 0.119 LPIPS, within
$0.07$\,dB, $0.0006$ and $0.001$ of every training-loop ablation
(supplement); the renderings shown are ordinary 3DGS quality for these
captures, which the method neither depends on nor improves.
Cost comparisons treat the reconstruction as sunk and compare only the
added cost of scale estimation; a user who needs only a scale is better
served by MASt3R end to end.

\paragraph{Reporting conventions.}
Unless stated otherwise, \emph{mean} and \emph{median} are deployed
errors $|\hat{s}/s^{*}-1|$ over \emph{all} scenes of a set, abstained
scenes contributing the warm estimate's error; ``on our 28'' is the fixed
set of ScanNet++ scenes the frozen estimator acts on. $p$-values are
paired Wilcoxon signed-rank tests; intervals are 95\% bootstrap CIs of
the paired difference.

\subsection{Main results: ScanNet++}
\label{sec:exp-snpp}
\begin{figure*}[t]
  \centering
  \includegraphics[width=0.43\linewidth]{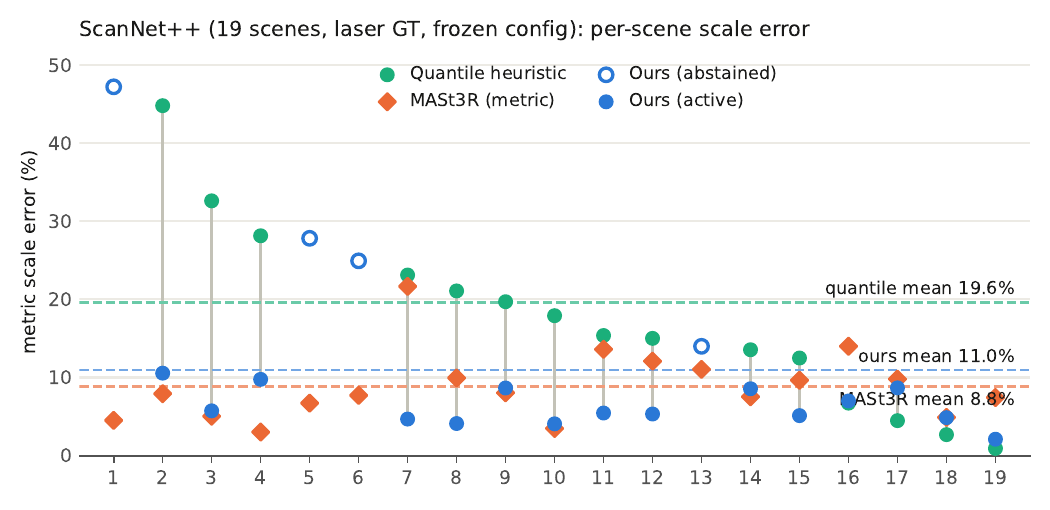}%
  \hspace{0.03\linewidth}%
  \includegraphics[width=0.225\linewidth]{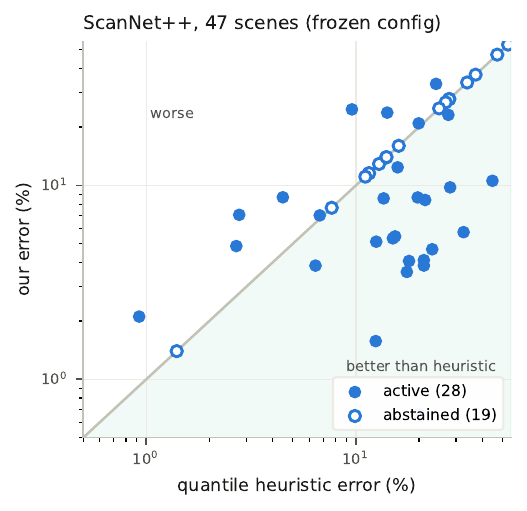}
  \caption{\textbf{Per-scene metric scale error on ScanNet++} (sorted by
  quantile error). Gray bars connect the quantile estimate to ours; open
  circles mark abstained scenes, which inherit the warm estimate. Left:
  the initial 19-scene set. Right: all 47 scenes against the identity
  line; active corrections mostly land below the diagonal, abstentions
  sit on it, and 9/47 end above. Left-panel indices map to scene IDs in
  the supplement.}
  \label{fig:snpp}
\end{figure*}
\begin{table}[t]
  \centering
  \caption{\textbf{Metric scale error on 47 ScanNet++ scenes} (laser-aligned
  GT; configuration frozen before any public evaluation, the last 28
  scenes held out). The estimator acts on 28 scenes and abstains on 19; 9 scenes end
  worse than the heuristic (3 by ${>}9$\,pp, worst $+15$\,pp). The router
  sends only the abstained scenes to MASt3R.
  The last column is one \emph{fixed} scene set on every row (the 28
  scenes our frozen estimator acts on), so it can be read vertically.
  $^{\dagger}$Non-frozen sensitivity study: post-hoc estimator re-run
  with a stronger segmenter, nothing else changed.
  $^{\ddagger}$ScanNet++ is in MoGe-2's training corpus, so this row is
  in-domain and not comparable with the rest (outside it MoGe-2 errs by
  28--35\% on our captures and 5.6\% on ETH3D).}
  \label{tab:snpp}
  \small
  \begin{tabular}{lccc}
    \toprule
    & mean & median & on our 28 \\
    \midrule
    Quantile heuristic & 21.1\% & 17.5\% & 16.8\% \\
    DA-3 alignment & 28.8\% & 22.5\% & 30.2\% \\
    MoGe-2 alignment$^{\ddagger}$ & 2.3\% & 1.9\% & 2.1\% \\
    MASt3R (metric)    & 9.9\% & 7.9\% & 10.2\% \\
    \textbf{Ours (frozen)} & 16.6\% & 10.6\% & \textbf{9.3\%} \\
    Ours (Swin-L teacher)$^{\dagger}$ & 15.7\% & 8.7\% & 7.4\% \\
    \textbf{Ours\,+\,MASt3R router} & \textbf{9.4\%} & \textbf{7.0\%} & \textbf{9.3\%} \\
    \midrule
    \multicolumn{4}{l}{\emph{real-SfM condition} (COLMAP from images, 14 scenes)} \\
    Quantile heuristic & 19.7\% & 13.7\% & 17.8\% \\
    \textbf{Ours (frozen)} & \textbf{16.8\%} & \textbf{7.8\%} & \textbf{12.0\%} \\
    \bottomrule
  \end{tabular}
\end{table}
Table~\ref{tab:snpp} and Fig.~\ref{fig:snpp} summarize the flagship
benchmark. Against the in-pipeline heuristic, our estimator reduces the
mean error from 21.1\% to 16.6\% (median $17.5\%\!\to\!10.6\%$; Wilcoxon
signed-rank $p{=}2.6{\times}10^{-3}$) and, on the 28 scenes where it acts, from 16.8\% to 9.3\%. The distributions are heavy-tailed, so comparative
claims rest on paired tests and we report the confidence interval of the
paired \emph{difference}:
the reduction against the heuristic is $4.5$\,pp (95\% CI $[1.9, 7.2]$)
over all 47 scenes and $7.5$\,pp ($[3.3, 11.6]$) on the 28 it acts on.
Single-image metric models under the alignment recipe split sharply.
Depth Anything 3 lands at 28.8\%, worse than the heuristic it would
replace. MoGe-2 reaches 2.3\%, ahead of
everything here --- but ScanNet++ is in its own training corpus (176K
frames, 4.6\% sampling weight), so those scenes are in-domain for it and
held out for us. Outside that mixture it errs by 34.8\% on our development scenes and
28.4\% on the in-the-wild captures of Sec.~\ref{sec:exp-custom}, against
our 7.7\% and 12.3\%, while on ETH3D it is the better estimator (5.6\%
against our 21.1\%; one of its four scenes is in regime).
MASt3R is more accurate overall (9.9\%; its reconstruction replaces the
whole pipeline and it never abstains). On
the scenes our estimator handles itself, the two are statistically
indistinguishable (9.3\% vs.\ 10.2\%, 14/28 per-scene wins with one
tie, $p{=}0.19$), at over 20$\times$ lower estimation cost and inside the
existing COLMAP/3DGS pipeline. Our gate also carries a confidence signal MASt3R does not:
routing exactly the abstained scenes to MASt3R matches running MASt3R
everywhere (9.4\% vs.\ 9.9\%; the margin is insignificant, paired
$\Delta{=}{-}0.5$\,pp, 95\% CI $[-2.9,+2.0]$, $p{=}0.36$,
TOST-equivalent at $\pm3$\,pp; supplement) at 60\%
fewer invocations. This routed configuration is the deployment we
advocate and the paper's strongest result: our estimator answers where
its evidence is sufficient, and its refusals identify the scenes worth a
heavyweight model's cost. The saving alone does not justify the system:
at ${\sim}4$\,min per abstained scene it is modest beside a 6--16\,min
reconstruction.
The fixed column is a set our own gate chose, but not simply the easy
scenes: MASt3R is marginally \emph{more} accurate on the scenes we
abstain from (9.4\%) than on those we act on (10.2\%). Under the real-SfM protocol (bottom rows; COLMAP on 16 scenes fixed
before evaluation, 14 passing the 20\,cm alignment gate) the same
pattern holds against genuine COLMAP ambiguity: the median error halves,
and the known partial-cue failure scene recurs.

\paragraph{Failure analysis.}
Nine active corrections end worse than the heuristic, three by more than
9\,pp. The supplement dissects the worst ($+15$\,pp): at the
\emph{true} scale nearly every object cue measures only 10--60\% of
its prior (no full door is ever seen, 41--50\%) while the room-height
anchor sits at 82\%; dozens of partial-object lower bounds outvote the
one correct anchor, and the gate cannot catch it because those partial
cues agree with each other (inlier mass 1.7--1.9). This is the principal open failure mode, the partial-instance
limitation stated in Sec.~\ref{sec:method-cues}, but only partly a
geometric one: swapping in a stronger segmenter (Swin-L, 57.3 vs.\ 47.2
mIoU), nothing else changed, repairs this scene ($24.6\%\!\to\!8.6\%$)
and cuts error from 9.7\% to 6.3\% on the 24 scenes both teachers act on
(a paired reduction of $3.5$\,pp, 95\% CI $[1.5, 5.9]$, $p{=}0.01$), so
much of what looks like partial \emph{reconstruction} is partial
\emph{labeling} (supplement). We report it as a sensitivity study, having
frozen the configuration before public evaluation.

\subsection{Held-out safety: pre-registered abstention}
\label{sec:exp-heldout}
Fifty additional ARKitScenes videos, the handheld-video regime in which
development-time studies showed the cue mass collapsing, were processed
with the frozen pipeline after we logged the prediction ``mostly
abstain, match the baseline.'' Outcome: 50/50 abstentions, estimate
equal to the heuristic on every video, zero degradations; ScanNet++
\emph{iPhone} video behaves the same (3/3).

\subsection{Development-set results}
\label{sec:exp-dev}
On the five development captures (real COLMAP, tape-measured heights),
the delivered room height errs by 27.4\% for the quantile heuristic and,
under an identical protocol, 66.3\%, 70.9\% and 34.8\% for metric-depth
alignment with Depth Anything 3, Depth Pro and MoGe-2~\cite{moge2}
(Fig.~\ref{fig:teaser}), versus 7.7\% for the full system. The
configuration was tuned here and one marginal scene is unstable across
re-runs ($\pm$20\,pp), which is why headline claims rest on the frozen
benchmarks.

\subsection{What the accuracy rests on}
\label{sec:exp-e5}
\textbf{The fusion buys robustness.} A gap-only or prior-mean-only estimator matches our median but
its \emph{mean} balloons to $\sim$40\%, because it has no defense where
the gap cue fails; full fusion cuts that mean by $2.4\times$. Accuracy
therefore does not come from the room-height prior alone. It is also insensitive to how the priors are set: shifting every prior
mean by $\pm10\%$ moves the mean only to 17.9--18.5\%, and perturbing
the anchors \emph{independently}, as a regional catalog would, is
cheaper still (200 draws: 17.1\% median, 18.7\% worst). Discarding the
reliability weights costs 31\,pp. \textbf{Visibility gating buys coverage, not accuracy.} Removing it leaves the deployed mean
unchanged (16.4\%, $p{=}0.82$) but pushes 12 more scenes below the
cue-mass threshold (31/47 abstain), cutting the set it can correct from
28 to 16 and raising the router's MASt3R bill from 19 calls to 31; the
symmetric $|z{-}D|$ test is strictly worse on every metric
($p{=}0.03$), since grazing-incidence culling starves floor cues. A softer contribution-weighted gate in the spirit of~\cite{vala,labelgs}
is also worse (17.6\%, 22 abstentions; supplement).
\textbf{The training loop delivers the scale; it does not improve
it.} Retraining all 47 scenes with the scale pinned at the warm start
and running the same frozen estimator gives 16.5\%/9.2\% against our
16.6\%/10.6\% (paired $0.2$\,pp, 95\% CI $[-1.8,+1.4]$, $p{=}0.28$), and
no single loop component moves the development split either (paired CIs
all cross zero, $p{\ge}0.23$). The accuracy is the fusion's; the loop
delivers a model that is already metric and keeps densification in
meters, since
forced training at $0.1\times$ the metric scale loses 0.7--4.1\,dB and
destabilizes the Gaussian budget by 31--40\%, while nominal to
$10\times$ sits at seed-noise level. The frozen thresholds are imperfect in hindsight:
$M_{\min}{=}1.25$ would have given 15.2\% against our 16.6\%. A
segmenter ten mIoU stronger peaks at the same threshold, so the gate
needs no per-teacher calibration (supplement).

\subsection{Operating regime and negative results}
\label{sec:exp-regime}
\begin{figure}[t]
  \centering
  \includegraphics[width=0.64\linewidth]{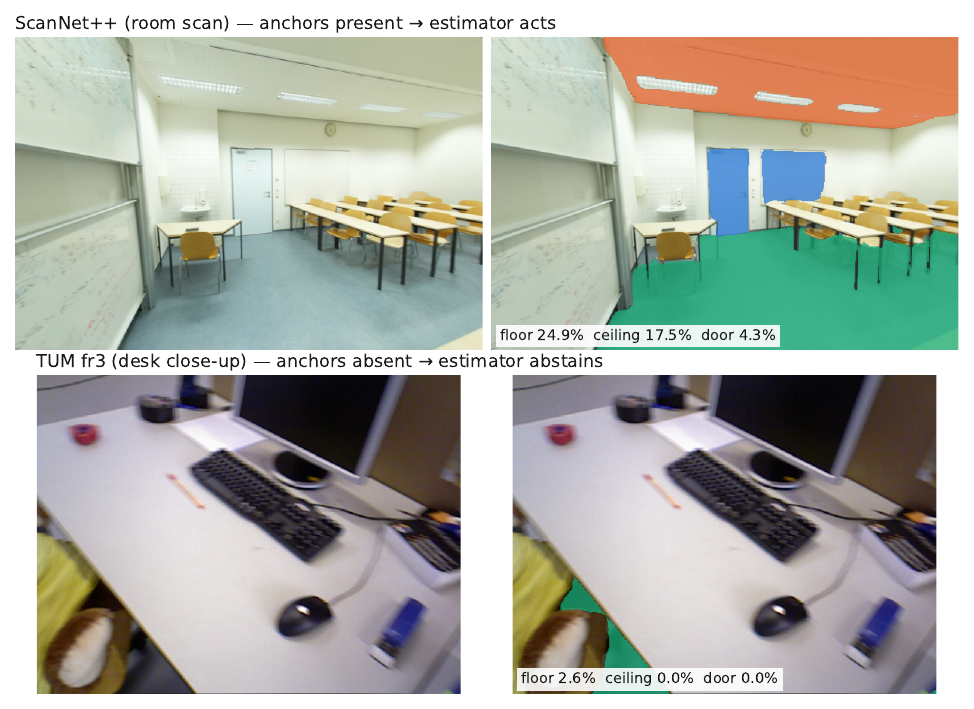}
  \caption{\textbf{Abstention tracks capture content across regimes.}
  Teacher segmentations of raw frames: a ScanNet++ room scan shows the
  anchor classes on a quarter of its pixels (top); a TUM desk sequence
  shows essentially none (bottom).}
  \label{fig:abstain-ev}
\end{figure}
The estimator's territory is real photographs of rooms whose architecture
is visible. Outside it, the system refuses predictably: sequences whose
images barely contain the anchor classes (TUM desk close-ups: 0.0\%
ceiling/door pixels, Fig.~\ref{fig:abstain-ev}); handheld video sweeps
(ARKitScenes: 50/50 abstain; ScanNet++ iPhone: 3/3); and synthetic
appearance, which breaks the frozen teacher outright (Replica).

\paragraph{A refuted explanation.}
We attributed the video failures to resolution and pre-registered that
re-running ten held-out videos at $640{\times}480$ (only the pixels
change) would convert some to active. It did not: $6.25\times$ the pixel
count leaves 10/10 abstentions, moving mean cue mass from 0.33 to 0.41
against the 1.5 gate. Labels survive (16--43 cues per scene) but the
measurements \emph{disagree}, scattering at a median $5.3\sigma$ with
only 20\% inside $2\sigma$: handheld sweeps fragment instances, and
fragmented cues contradict each other. This unifies the two failure
modes: when partial cues \emph{agree} we get the confident error of the
failure analysis, and when they disagree the gate fires. The operative
variable is cue disagreement, not resolution.

\paragraph{Two levels of evidence.}
ETH3D probes the boundary: one of its four scenes is in regime and the
estimator acts there ($16.5\%\!\to\!9.1\%$), abstaining on the outdoor
and industrial ones (supplement). Abstention evidence
is two-level: \emph{between} regimes the teacher's 2D anchor coverage
separates acting from abstaining; \emph{within} ScanNet++ it does not
($p{=}0.96$). That second level is why coverage cannot \emph{replace}
the gate: routing by coverage instead of cue mass yields 14.5--15.3\% at
the same 19-call budget against the gate's 9.4\%, and needs 37--47 of the
47 scenes sent out to reach 9.6\%, by which point the saving is gone.

\subsection{Deployment on captures of our own}
\label{sec:exp-custom}
\begin{table}[t]
  \centering
  \caption{\textbf{Frozen system on 13 captures we recorded ourselves}
  (phones and a drone; no tuning, no per-dataset configuration).
  Delivered floor$\to$ceiling height in meters, signed error against a
  nominal per-venue height. Captures with no floor$\to$ceiling cue have
  no delivered height. MoGe-2: identical protocol and operator. MASt3R
  fails the 20\,cm registration gate on 12 of 13 captures (supplement).}
  \label{tab:custom}
  \scriptsize
  \setlength{\tabcolsep}{1.2pt}
  \begin{tabular}{lccccccc}
\toprule
capture & nom. & quantile & MoGe-2 & ours & mass & cues & decision \\
\midrule
Apartment A, session 1 & 2.3 & 2.90\,(+26) & 1.77\,(-23) & 2.65\,(+15) & 2.03 & 45 & act \\
Apartment A, session 2 & 2.3 & 3.03\,(+32) & 2.40\,(+4) & 2.36\,(+3) & 1.89 & 57 & act \\
Apartment A, session 3 & 2.3 & 2.75\,(+20) & 3.48\,(+52) & 2.75\,(+20) & 0.73 & 21 & abstain \\
Home B, phone 1 & 2.3 & --- & --- & --- & 0.98 & 33 & abstain \\
Home B, phone 2 & 2.3 & 2.62\,(+14) & 2.04\,(-11) & 2.62\,(+14) & 0.84 & 30 & abstain \\
Home C, stills & 2.3 & 1.91\,(-17) & 1.34\,(-42) & 1.91\,(-17) & 0.52 & 37 & abstain \\
Home C, video & 2.3 & 3.08\,(+34) & 2.04\,(-11) & 2.67\,(+16) & 1.53 & 39 & act \\
Show flat & 2.3 & 2.63\,(+14) & 2.22\,(-3) & 2.63\,(+14) & 1.46 & 39 & abstain \\
House corridor & 2.3 & 1.24\,(-46) & 0.70\,(-70) & 2.12\,(-8) & 2.40 & 39 & act \\
Robotics lab & 3.0 & --- & --- & --- & 0.03 & 9 & abstain \\
Office floor & 3.0 & 2.29\,(-24) & 1.82\,(-39) & 3.12\,(+4) & 1.50 & 28 & act \\
Aerial (drone) & --- & --- & --- & --- & 0.00 & 0 & abstain \\
Sports field & --- & --- & --- & --- & 0.00 & 6 & abstain \\
\bottomrule
\end{tabular}

\end{table}
\begin{figure*}[t]
  \centering
  \includegraphics[width=0.82\linewidth]{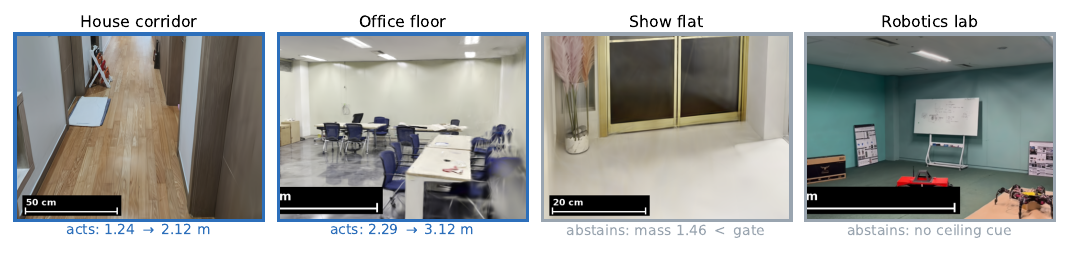}
  \caption{\textbf{What the frozen system does on captures of our own.}
  Renderings from four of the 13 reconstructions, each with a reference
  bar at the delivered metric scale. Blue: the estimator acts and corrects
  the ceiling height. Gray: it refuses --- the show flat's cues disagree
  just below the gate, the lab capture never frames a ceiling --- and
  returns the warm estimate.}
  \label{fig:custom-qual}
  \vspace{-8pt}
\end{figure*}
To see what happens on captures nobody designed the system around, we
recorded 13 of our own with consumer phones and a drone
(Tab.~\ref{tab:custom}), reconstructed from images alone by COLMAP so
that no inertial signal survives into the reconstruction, the regime
Sec.~\ref{sec:intro} identifies. Nothing else changed: same nominal
height, thresholds, teacher and seeds as the ScanNet++ runs. Ground
truth is weaker here and we say so: the reference is a nominal ceiling
height per venue class ($2.3$\,m for dwellings, $3.0$\,m for
institutional interiors), so it sets a floor on the error any method can
show and these percentages are not comparable with the laser-referenced
ones; we read the study as a transfer test of the frozen configuration,
not as a second accuracy result. It does not decide the comparison,
though: over an assumed true dwelling height of $2.1$--$2.7$\,m our mean
error stays below MoGe-2's throughout ($19.6\%$ vs.\ $26.3\%$ at
worst), with $2.3$\,m nearer our unfavorable end (supplement).

The behavior transfers (Fig.~\ref{fig:custom-qual}; per-capture detail
in the supplement). The estimator acts on five captures and improves
every one, and the largest corrections are where the heuristic is worst:
a corridor whose vertical span is truncated ($-45.9\%\!\to\!-7.9\%$) and
an office floor where the quantile heuristic imposes a $2.5$\,m room on a
$3.0$\,m hall ($-23.8\%\!\to\!+4.1\%$). That last correction runs
\emph{away} from the nominal prior, so the measured cues and not the
assumed height carry it. Eight captures abstain and inherit their warm
baseline exactly; none degrades. The regime boundary holds on data that
never informed it: the sports field and the aerial survey abstain at
mass $0.00$.

\paragraph{What the error means in practice.}
\label{sec:exp-apps}
A $1.70$\,m mannequin composited at the heuristic's scale becomes a
$+48\%$ giant where ours is visually correct, and a door measured in the
model errs by 21\,cm at the median under our scale versus 35\,cm under
the heuristic (supplement). A global scale presumes no spatial drift:
over our longest real-SfM trajectory (703 frames) the per-region
floor-to-ceiling gap varies by only 3\% (CV).

\section{Limitations}
\label{sec:limitations}
Our cue set presumes \emph{indoor} scenes photographed as \emph{real}
images, with hand-specified size priors (regional catalogs:
supplement): synthetic appearance
breaks the frozen teacher, and captures that never show architecture
leave nothing to fuse. Abstention turns these into explicit refusals,
though it then returns the warm estimate unchanged, and only the no-anchor case is predictable in advance:
handheld video has anchors and still fails, on cue disagreement visible
only after measurement (Sec.~\ref{sec:exp-regime}). Partial instances
remain the open technical limitation (Sec.~\ref{sec:method-cues}): a
completeness filter converts degradations into refusals (9 to 4
degraded scenes, 12 more abstentions) but, held out, leaves deployed
error where it was ($p{=}0.75$; supplement). The assumed room height is
the one sensitive input: retraining all 47 scenes at 2.1 and 3.0\,m
gives 13.8\% and 25.3\% against 16.6\% at 2.5\,m, all below the
heuristic (supplement). Accuracy is bounded by the
teacher, which is also an upgrade path: a stronger segmenter cuts error
by a third at the same gate threshold (Sec.~\ref{sec:exp-snpp}). The positive evidence is concentrated on captures of
our own because no public indoor benchmark we know of is both in regime and outside this generation's training data: ScanNet++
sits inside MoGe-2's mixture, ETH3D offers one in-regime scene, Replica
is synthetic, TUM shows no architecture, and ARKitScenes carries the
inertial signal that makes the problem moot.

\section{Conclusion}
\label{sec:conclusion}
We presented a selective metric-scale estimator for indoor 3DGS: it
reads the scale off the reconstruction's own semantic content,
visibility-aware voting lifts the scenes it can act on from 16 to 28 (a
coverage gain at unchanged accuracy), and an inlier-mass test turns weak
evidence into a refusal rather than a confident wrong answer. Frozen
before first contact, it cuts the heuristic's mean error by a fifth;
its strongest result is the selective hybrid, which routes its refusals
to match MASt3R at 60\% fewer calls. On benchmark-grade imagery a
strong feed-forward model is the better estimator; our case rests on ordinary captures, where ours is more
reliable and says when it is not confident. The same model scores $2.3\%$ inside its training mixture and
$28$--$35\%$ on unseen captures: a published number is a best case, not
a deployment estimate.

\paragraph{Acknowledgments.}
This work was supported by Institute for Information \& communications
Technology Promotion (IITP) grant funded by the Korea government (MSIT)
(No.~RS-2026-25522152).
{
    \small
    \bibliographystyle{ieeenat_fullname}
    \bibliography{main}

@String(CVPR= {IEEE Conf. Comput. Vis. Pattern Recog.})

@String(ICCV= {Int. Conf. Comput. Vis.})

@String(ECCV= {Eur. Conf. Comput. Vis.})

@String(ICLR = {Int. Conf. Learn. Represent.})

@String(CVPR  = {CVPR})

@String(ICCV  = {ICCV})

@String(ECCV  = {ECCV})

@String(ICLR  = {ICLR})

@article{kerbl3Dgaussians,
  author  = {Kerbl, Bernhard and Kopanas, Georgios and Leimk{\"u}hler, Thomas and Drettakis, George},
  title   = {3D Gaussian Splatting for Real-Time Radiance Field Rendering},
  journal = {ACM Transactions on Graphics},
  volume  = {42},
  number  = {4},
  year    = {2023}
}

@inproceedings{schoenberger2016sfm, author={Sch\"onberger, Johannes L. and Frahm, Jan-Michael}, title={Structure-from-Motion Revisited}, booktitle={CVPR}, year={2016}}

@inproceedings{mipsplatting, author={Yu, Zehao and Chen, Anpei and Huang, Binbin and Sattler, Torsten and Geiger, Andreas}, title={Mip-Splatting: Alias-free 3D Gaussian Splatting}, booktitle={CVPR}, year={2024}}

@inproceedings{taming3dgs, author={Mallick, Saswat and Goel, Rahul and Kerbl, Bernhard and Steinberger, Markus and Carrasco, Francisco Vicente and De La Torre, Fernando}, title={Taming 3DGS: High-Quality Radiance Fields with Limited Resources}, booktitle={SIGGRAPH Asia}, year={2024}}

@article{depthanythingv2, author={Yang, Lihe and Kang, Bingyi and Huang, Zilong and Zhao, Zhen and Xu, Xiaogang and Feng, Jiashi and Zhao, Hengshuang}, title={Depth Anything V2}, journal={NeurIPS}, year={2024}}

@article{depthanything3, author={Lin, Haotong and Chen, Sili and Liew, Junhao and Chen, Donny Y. and Li, Zhenyu and Shi, Guang and Feng, Jiashi and Kang, Bingyi}, title={Depth Anything 3: Recovering the Visual Space from Any Views}, journal={arXiv:2511.10647}, year={2025}}

@inproceedings{metric3d, author={Yin, Wei and Zhang, Chi and Chen, Hao and Cai, Zhipeng and Yu, Gang and Wang, Kaixuan and Chen, Xiaozhi and Shen, Chunhua}, title={Metric3D: Towards Zero-shot Metric 3D Prediction from a Single Image}, booktitle={ICCV}, year={2023}}

@article{zoedepth, author={Bhat, Shariq Farooq and Birkl, Reiner and Wofk, Diana and Wonka, Peter and M\"uller, Matthias}, title={ZoeDepth: Zero-shot Transfer by Combining Relative and Metric Depth}, journal={arXiv:2302.12288}, year={2023}}

@article{vipe, author={Huang, Jiahui and Zhou, Qunjie and Rabeti, Hesam and Korovko, Aleksandr and Ling, Huan and Ren, Xuanchi and Shen, Tianchang and Gao, Jun and Slepichev, Dmitry and Lin, Chen-Hsuan and Ren, Jiawei and Xie, Kevin and Biswas, Joydeep and Leal-Taix{\'e}, Laura and Fidler, Sanja}, title={ViPE: Video Pose Engine for 3D Geometric Perception}, journal={arXiv:2508.10934}, year={2025}}

@inproceedings{dust3r, author={Wang, Shuzhe and Leroy, Vincent and Cabon, Yohann and Chidlovskii, Boris and Revaud, Jerome}, title={DUSt3R: Geometric 3D Vision Made Easy}, booktitle={CVPR}, year={2024}}

@inproceedings{mast3r, author={Leroy, Vincent and Cabon, Yohann and Revaud, Jerome}, title={Grounding Image Matching in 3D with MASt3R}, booktitle={ECCV}, year={2024}}

@inproceedings{vggt, author={Wang, Jianyuan and Chen, Minghao and Karaev, Nikita and Vedaldi, Andrea and Rupprecht, Christian and Novotny, David}, title={VGGT: Visual Geometry Grounded Transformer}, booktitle={CVPR}, year={2025}}

@article{scaleSLAM, author={Frost, Duncan and Prisacariu, Victor and Murray, David}, title={Recovering Stable Scale in Monocular SLAM Using Object-Supplemented Bundle Adjustment}, journal={IEEE Transactions on Robotics}, year={2018}}

@inproceedings{objectSLAM, author={Salas-Moreno, Renato F. and Newcombe, Richard A. and Strasdat, Hauke and Kelly, Paul H. J. and Davison, Andrew J.}, title={SLAM++: Simultaneous Localisation and Mapping at the Level of Objects}, booktitle={CVPR}, year={2013}}

@inproceedings{feature3dgs, author={Zhou, Shijie and Chang, Haoran and Jiang, Sicheng and Fan, Zhiwen and Zhu, Zehao and Xu, Dejia and Chari, Pradyumna and You, Suya and Wang, Zhangyang and Kadambi, Achuta}, title={Feature 3DGS: Supercharging 3D Gaussian Splatting to Enable Distilled Feature Fields}, booktitle={CVPR}, year={2024}}

@inproceedings{langsplat, author={Qin, Minghan and Li, Wanhua and Zhou, Jiawei and Wang, Haoqian and Pfister, Hanspeter}, title={LangSplat: 3D Language Gaussian Splatting}, booktitle={CVPR}, year={2024}}

@article{planargs, author={Jin, Xirui and Jin, Renbiao and Li, Boying and Zou, Danping and Yu, Wenxian}, title={PlanarGS: High-Fidelity Indoor 3D Gaussian Splatting Guided by Vision-Language Planar Priors}, journal={arXiv:2510.23930}, year={2025}}

@article{gsroom, author={Zhang, Wanting and Xiang, Haodong and Liao, Zhichao and Lai, Xiansong and Li, Xinghui and Zeng, Long}, title={2DGS-Room: Seed-Guided 2D Gaussian Splatting with Geometric Constraints for High-Fidelity Indoor Scene Reconstruction}, journal={arXiv:2412.03428}, year={2024}}

@inproceedings{spcgs, author={Liao, Guibiao and Li, Qing and Bao, Zhenyu and Qiu, Guoping and Liu, Kanglin}, title={SPC-GS: Gaussian Splatting with Semantic-Prompt Consistency for Indoor Open-World Free-view Synthesis from Sparse Inputs}, booktitle={CVPR}, year={2025}}

@inproceedings{mask2former, author={Cheng, Bowen and Misra, Ishan and Schwing, Alexander G. and Kirillov, Alexander and Girdhar, Rohit}, title={Masked-attention Mask Transformer for Universal Image Segmentation}, booktitle={CVPR}, year={2022}}

@inproceedings{scannetpp, author={Yeshwanth, Chandan and Liu, Yueh-Cheng and Nie{\ss}ner, Matthias and Dai, Angela}, title={ScanNet++: A High-Fidelity Dataset of 3D Indoor Scenes}, booktitle={ICCV}, year={2023}}

@inproceedings{arkitscenes, author={Baruch, Gilad and Chen, Zhuoyuan and Dehghan, Afshin and Dimry, Tal and Feigin, Yuri and Fu, Peter and Gebauer, Thomas and Joffe, Brandon and Kurz, Daniel and Schwartz, Arik and Shulman, Elad}, title={ARKitScenes: A Diverse Real-World Dataset for 3D Indoor Scene Understanding Using Mobile RGB-D Data}, booktitle={NeurIPS Datasets}, year={2021}}

@inproceedings{eth3d, author={Sch\"ops, Thomas and Sch\"onberger, Johannes L. and Galliani, Silvano and Sattler, Torsten and Schindler, Konrad and Pollefeys, Marc and Geiger, Andreas}, title={A Multi-View Stereo Benchmark with High-Resolution Images and Multi-Camera Videos}, booktitle={CVPR}, year={2017}}

@inproceedings{tum, author={Sturm, J\"urgen and Engelhard, Nikolas and Endres, Felix and Burgard, Wolfram and Cremers, Daniel}, title={A Benchmark for the Evaluation of RGB-D SLAM Systems}, booktitle={IROS}, year={2012}}

@inproceedings{replica, author={Straub, Julian and Whelan, Thomas and Ma, Lingni and Chen, Yufan and Wijmans, Erik and Green, Simon and Engel, Jakob J. and Mur-Artal, Raul and Ren, Carl and Verma, Shobhit and Clarkson, Anton and Yan, Mingfei and Budge, Brian and Yan, Yajie and Pan, Xiaqing and Yon, June and Zou, Yuyang and Leon, Kimberly and Carter, Nigel and Briales, Jesus and Gillingham, Tyler and Mueggler, Elias and Pesqueira, Luis and Savva, Manolis and Batra, Dhruv and Strasdat, Hauke M. and De Nardi, Renzo and Goesele, Michael and Lovegrove, Steven and Newcombe, Richard}, title={The Replica Dataset: A Digital Replica of Indoor Spaces}, booktitle={arXiv:1906.05797}, year={2019}}

@article{mapanything, author={Keetha, Nikhil and M{\"u}ller, Norman and Sch{\"o}nberger, Johannes and Porzi, Lorenzo and Zhang, Yuchen and Fischer, Tobias and Knapitsch, Arno and Zauss, Duncan and Weber, Ethan and Antunes, Nelson and Luiten, Jonathon and Lopez-Antequera, Manuel and Rota Bul{\`o}, Samuel and Richardt, Christian and Ramanan, Deva and Scherer, Sebastian and Kontschieder, Peter}, title={MapAnything: Universal Feed-Forward Metric 3D Reconstruction}, journal={arXiv:2509.13414}, year={2025}}

@article{metric3dv2, author={Hu, Mu and Yin, Wei and Zhang, Chi and Cai, Zhipeng and Long, Xiaoxiao and Wang, Kaixuan and Chen, Hao and Yu, Gang and Shen, Chunhua and Shen, Shaojie}, title={Metric3D v2: A Versatile Monocular Geometric Foundation Model for Zero-shot Metric Depth and Surface Normal Estimation}, journal={IEEE TPAMI}, year={2024}}

@inproceedings{unidepth, author={Piccinelli, Luigi and Yang, Yung-Hsu and Sakaridis, Christos and Segu, Mattia and Li, Siyuan and Van Gool, Luc and Yu, Fisher}, title={UniDepth: Universal Monocular Metric Depth Estimation}, booktitle={CVPR}, year={2024}}

@inproceedings{depthpro, author={Bochkovskii, Aleksei and Delaunoy, Ama{\"e}l and Germain, Hugo and Santos, Marcel and Zhou, Yichao and Richter, Stephan R. and Koltun, Vladlen}, title={Depth Pro: Sharp Monocular Metric Depth in Less Than a Second}, booktitle={ICLR}, year={2025}}

@inproceedings{labelgs, author={Zhang, Yupeng and Zheng, Dezhi and Lu, Ping and Zhang, Han and Wang, Lei and Xiang, Liping and Luo, Cheng and Deng, Kaijun and Fu, Xiaowen and Shen, Linlin and Wang, Jinbao}, title={LabelGS: Label-Aware 3D Gaussian Splatting for 3D Scene Segmentation}, booktitle={PRCV}, year={2025}}

@inproceedings{moge2, author={Wang, Ruicheng and Xu, Sicheng and Dong, Yue and Deng, Yu and Xiang, Jianfeng and Lv, Zelong and Sun, Guangzhong and Tong, Xin and Yang, Jiaolong}, title={MoGe-2: Accurate Monocular Geometry with Metric Scale and Sharp Details}, booktitle={NeurIPS}, year={2025}}

@article{amb3r, author={Wang, Hengyi and Agapito, Lourdes}, title={AMB3R: Accurate Feed-forward Metric-scale 3D Reconstruction with Backend}, journal={arXiv:2511.20343}, year={2025}}

@inproceedings{vala, author={Wang, Sen and Li, Kunyi and Liang, Siyun and Alegret, Elena and Ma, Jing and Navab, Nassir and Gasperini, Stefano}, title={Visibility-Aware Language Aggregation for Open-Vocabulary Segmentation in 3D Gaussian Splatting}, booktitle={3DV}, year={2026}}

@article{vgpnav, author={Pan, Hewei and Zhu, Weiye and Zhang, Zekai and Huang, Zitong and Xu, Rongtao and Wang, Jinbao and Zheng, Feng}, title={VGP-Nav: Metric-Aware Visual Geometric Perception for Robot Navigation}, journal={arXiv:2606.09268}, year={2026}}

@article{hislam2, author={Zhang, Wei and Cheng, Qing and Skuddis, David and Zeller, Niclas and Cremers, Daniel and Haala, Norbert}, title={HI-SLAM2: Geometry-Aware Gaussian SLAM for Fast Monocular Scene Reconstruction}, journal={IEEE Transactions on Robotics}, volume={41}, pages={6478--6493}, year={2025}}

@inproceedings{pi3, author={Wang, Yifan and Zhou, Jianjun and Zhu, Haoyi and Chang, Wenzheng and Zhou, Yang and Li, Zizun and Chen, Junyi and Pang, Jiangmiao and Shen, Chunhua and He, Tong}, title={$\pi^3$: Permutation-Equivariant Visual Geometry Learning}, booktitle={ICLR}, year={2026}}
}

\clearpage
\appendix
\setcounter{table}{0}\renewcommand{\thetable}{S\arabic{table}}
\setcounter{figure}{0}\renewcommand{\thefigure}{S\arabic{figure}}
\setcounter{equation}{0}\renewcommand{\theequation}{S\arabic{equation}}
\twocolumn[%
  \begin{center}
    {\Large\bfseries Supplementary Material}\par\vspace{0.5em}
    {\large METRIC-GS: Selective Metric-Scale Estimation for 3D Gaussian Splatting}\par
  \end{center}
  \vspace{1.5em}
]

\section*{Where each claim is substantiated}
Every comparative claim in the main paper is backed by a paired test and
a stated protocol; this index maps each claim to its evidence so that a
reviewer question can be answered from the submission itself.

\begin{table}[t]
  \centering\scriptsize
  \setlength{\tabcolsep}{3pt}
  \begin{tabular}{p{0.55\linewidth}p{0.37\linewidth}}
    \toprule
    claim (main paper) & evidence \\
    \midrule
    Frozen system: 16.6\%/10.6\% on 47 laser-GT ScanNet++ scenes; acts on 28, abstains on 19 & Tab.~1; per scene supp.~Tab.~\ref{tab:supp-scenes} \\
    Router parity: routing only abstentions to MASt3R matches MASt3R everywhere ($-0.5$\,pp, n.s.; TOST-equivalent at $\pm3$\,pp) & \S4.2; supp.~\S\ref{supp:stats}, Tabs.~\ref{tab:supp-covroute}--\ref{tab:supp-ci} \\
    Prior sensitivity: uniform $\pm10\%$ shifts, heterogeneous and regional catalogs, MC worst case & \S4.5; supp.~\S\ref{supp:locale}, Tabs.~\ref{tab:supp-perturb}--\ref{tab:supp-locale} \\
    Abstention threshold: frozen $M_{\min}{=}1.5$ on the mean plateau; $1.25$ would score 15.2\% (reported) & \S4.5; supp.~Tab.~\ref{tab:supp-tau} \\
    Gate needs no per-teacher recalibration (optimum static under a ${+}10$ mIoU teacher) & \S4.5; supp.~\S\ref{supp:gate-teacher}, Tab.~\ref{tab:supp-gate-teacher} \\
    Visibility gate: one-sided culling; removal or softening costs actionability or accuracy & \S4.5; supp.~\S\ref{supp:softvis}, Tab.~\ref{tab:supp-vis} \\
    In-loop scale is delivery, not accuracy: post-hoc-only retrain 16.5\% vs.\ 16.6\% ($p{=}0.28$) & \S4.5 (47-scene retrain, aggregate only); per-component ablation on the 19-scene split: supp.~\S\ref{supp:ablation19}, Tabs.~\ref{tab:supp-abl19-agg}--\ref{tab:supp-abl19} \\
    Teacher bound: a stronger segmenter alone reaches 6.3\% on the both-act subset (significant) & Tab.~1 ($\dagger$); supp.~\S\ref{supp:teacher}, Tab.~\ref{tab:supp-teacher} \\
    Held-out abstention: ARKitScenes at two resolutions; ScanNet++ iPhone 3/3; zero degradations & \S4.3; supp.~\S\ref{supp:ladder}, Tab.~\ref{tab:supp-ladder} \\
    Nominal-GT sensitivity: capture conclusions invariant to the assumed per-venue height & supp.~\S\ref{supp:custom}, Tab.~\ref{tab:supp-gtsens} \\
    Benchmark contamination: MoGe-2 2.3\% inside its training mixture, 28--35\% on unseen captures & \S4.2; supp.~\S\ref{supp:indomain}, Tabs.~\ref{tab:supp-ood}--\ref{tab:supp-indomain} \\
    In-the-wild captures: acts on 5/13 and improves every one; abstentions inherit the warm baseline, none degrades & \S4.7, Tab.~2; supp.~\S\ref{supp:custom} \\
    Rendering quality neutral on average (mean $\Delta$PSNR within $0.07$\,dB), per-scene moves up to $1.3$\,dB both ways & supp.~\S\ref{supp:quality}, Tab.~\ref{tab:supp-quality} \\
    Cost: ${\sim}11$\,s post-hoc estimation (the MLE itself ${<}10$\,ms); router saves 60\% of MASt3R invocations & \S4.2; supp.~\S\ref{supp:cost}, Tab.~\ref{tab:supp-cost} \\
    Negative results, kept: render-to-measure, asymmetric loss, soft visibility, completeness filter, scoped directions & supp.~\S\S\ref{supp:r2m}, \ref{supp:asym}, \ref{supp:softvis}, \ref{supp:completeness}, \ref{supp:scoped} \\
    \bottomrule
  \end{tabular}
\end{table}

\section{Per-scene results}
\label{supp:per-scene}
Table~\ref{tab:supp-scenes} lists the per-scene metric scale errors behind
the main table. Scene \texttt{5f99900f09} is excluded: its densification
exhausts a 16\,GB GPU at full and half resolution; we report this rather
than silently dropping it. Table~\ref{tab:supp-figidx} maps the scene
indices of the main paper's Fig.~3 left panel to their ScanNet++ IDs.

\begin{table}[htbp]
  \centering\scriptsize
  \caption{Scene indices of the main paper's Fig.~3 left panel (the
  initial 19-scene set, sorted by quantile error).}
  \label{tab:supp-figidx}
  \setlength{\tabcolsep}{2pt}
  \begin{tabular}{clclclcl}
    \toprule
    \# & scene & \# & scene & \# & scene & \# & scene \\
    \midrule
    1 & \texttt{40aec5fffa} & 6 & \texttt{fb5a96b1a2} & 11 & \texttt{a24f64f7fb} & 16 & \texttt{3e8bba0176} \\
    2 & \texttt{38d58a7a31} & 7 & \texttt{7831862f02} & 12 & \texttt{1ada7a0617} & 17 & \texttt{5eb31827b7} \\
    3 & \texttt{bde1e479ad} & 8 & \texttt{acd95847c5} & 13 & \texttt{f9f95681fd} & 18 & \texttt{cc5237fd77} \\
    4 & \texttt{21d970d8de} & 9 & \texttt{c50d2d1d42} & 14 & \texttt{5ee7c22ba0} & 19 & \texttt{3f15a9266d} \\
    5 & \texttt{c4c04e6d6c} & 10 & \texttt{5748ce6f01} & 15 & \texttt{7b6477cb95} & & \\
    \bottomrule
  \end{tabular}
\end{table}

\begin{table}[htbp]
  \centering\scriptsize
  \caption{ScanNet++ per-scene scale error (\%).}
  \label{tab:supp-scenes}
  \begin{tabular}{lcccc}
    \toprule
    scene & quantile & MASt3R & ours & decision \\
    \midrule
09c1414f1b & 2.8 & 25.3 & 7.0 & active \\
0d2ee665be & 9.6 & 9.2 & 24.6 & active \\
13c3e046d7 & 13.9 & 13.8 & 13.9 & abstain \\
1ada7a0617 & 15.0 & 12.1 & 5.3 & active \\
21d970d8de & 28.1 & 3.0 & 9.8 & active \\
25f3b7a318 & 14.1 & 35.3 & 23.7 & active \\
27dd4da69e & 12.9 & 4.6 & 12.9 & abstain \\
286b55a2bf & 37.2 & 5.7 & 37.2 & abstain \\
31a2c91c43 & 7.7 & 15.9 & 7.7 & abstain \\
38d58a7a31 & 44.8 & 7.9 & 10.6 & active \\
3db0a1c8f3 & 24.1 & 2.4 & 33.4 & active \\
3e8bba0176 & 6.7 & 14.0 & 7.0 & active \\
3f15a9266d & 0.9 & 7.4 & 2.1 & active \\
40aec5fffa & 47.2 & 4.5 & 47.2 & abstain \\
45b0dac5e3 & 11.5 & 7.4 & 11.5 & abstain \\
5748ce6f01 & 17.9 & 3.5 & 4.1 & active \\
578511c8a9 & 19.9 & 16.7 & 20.9 & active \\
5eb31827b7 & 4.5 & 9.8 & 8.7 & active \\
5ee7c22ba0 & 13.6 & 7.5 & 8.6 & active \\
6115eddb86 & 6.4 & 9.9 & 3.9 & active \\
7831862f02 & 23.1 & 21.7 & 4.7 & active \\
7b6477cb95 & 12.5 & 9.7 & 5.1 & active \\
7bc286c1b6 & 71.6 & 1.0 & 71.6 & abstain \\
825d228aec & 20.1 & 6.8 & 0.7 & active \\
9071e139d9 & 12.4 & 22.4 & 1.6 & active \\
99fa5c25e1 & 70.0 & 4.2 & 70.0 & abstain \\
a24f64f7fb & 15.4 & 13.6 & 5.5 & active \\
a8bf42d646 & 1.4 & 25.6 & 1.4 & abstain \\
a980334473 & 27.9 & 2.5 & 27.9 & abstain \\
ac48a9b736 & 21.4 & 14.4 & 8.4 & active \\
acd95847c5 & 21.1 & 9.9 & 4.1 & active \\
b0a08200c9 & 53.2 & 3.1 & 53.2 & abstain \\
bcd2436daf & 15.8 & 1.5 & 12.4 & active \\
bde1e479ad & 32.6 & 5.0 & 5.7 & active \\
c49a8c6cff & 27.6 & 0.8 & 23.1 & active \\
c4c04e6d6c & 27.8 & 6.7 & 27.8 & abstain \\
c50d2d1d42 & 19.7 & 8.0 & 8.7 & active \\
c5439f4607 & 26.8 & 18.2 & 26.8 & abstain \\
cc5237fd77 & 2.7 & 4.9 & 4.9 & active \\
d755b3d9d8 & 17.5 & 3.5 & 3.6 & active \\
e398684d27 & 11.6 & 11.8 & 11.6 & abstain \\
e7af285f7d & 16.0 & 2.4 & 16.0 & abstain \\
f2dc06b1d2 & 33.9 & 8.0 & 33.9 & abstain \\
f3685d06a9 & 11.1 & 24.8 & 11.1 & abstain \\
f3d64c30f8 & 21.1 & 0.3 & 3.9 & active \\
f9f95681fd & 14.0 & 11.0 & 14.0 & abstain \\
fb5a96b1a2 & 24.9 & 7.7 & 24.9 & abstain \\
    \bottomrule
  \end{tabular}
\end{table}

\section{Negative result: fronto-parallel render-to-measure}
\label{supp:r2m}
\begin{figure}[htbp]
  \centering
  \includegraphics[width=0.32\linewidth]{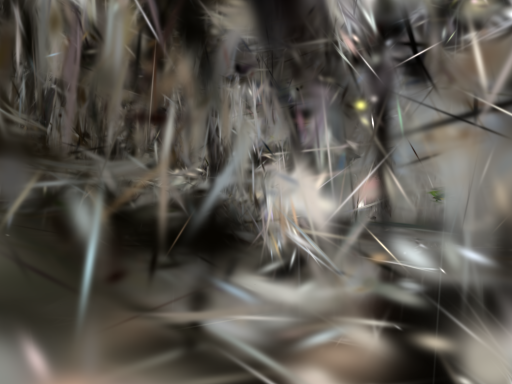}\hfill
  \includegraphics[width=0.32\linewidth]{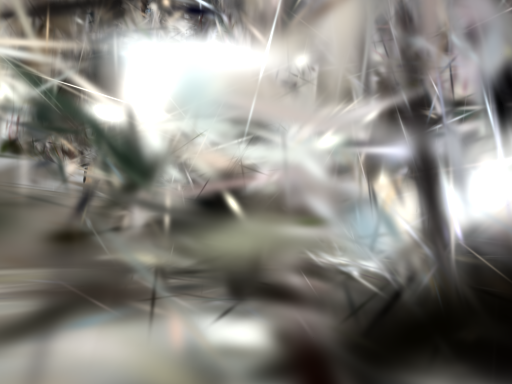}\hfill
  \includegraphics[width=0.32\linewidth]{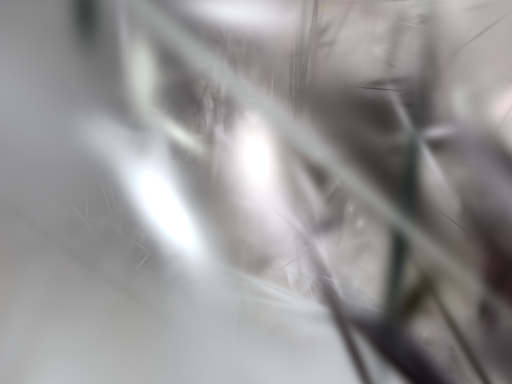}
  \caption{Fronto-parallel virtual views of door/wardrobe instances,
  including a view-manifold-constrained attempt (right): the renders
  degrade into floater fields away from the training trajectory, and
  depth-coverage gates cannot reject them because floaters occupy
  plausible depths.}
  \label{fig:supp-r2m}
\end{figure}

Synthesizing a fronto-parallel view of each detected instance and
re-measuring it with the teacher is appealing: the reconstruction can
"re-photograph" objects the capture only saw obliquely. In practice the
virtual viewpoints fall off the training-view manifold and the renders
collapse into floater fields (Fig.~\ref{fig:supp-r2m}); constraining the
camera toward the nearest training view and gating on depth coverage and
mask truncation does not rescue it, because floaters occupy plausible
depths. The surviving form of render-based supervision is the
visibility-aware voting of the main paper.

\section{Priors and hyperparameters}
Table~\ref{tab:supp-priors} lists every size prior; means and standard
deviations follow public architectural standards and typical furniture
dimensions and were never tuned per dataset; weights encode how
standardized a class is (doors and room height highest). The rules below
are the complete specification; with the table, they determine every
number in the paper.

\paragraph{Voting.} Every 200 training iterations over 24 randomly
selected training views (48 views and three seed ensembles
$\{12345,23456,34567\}$ for the reported post-hoc estimate; view subsets
come from fixed generators). A view votes for a Gaussian only if the
projection is in frame, camera depth $z>0.2$, and $z < D_v(1+0.05)$ for
the rendered depth $D_v$. A Gaussian keeps the winning class if it holds
$\ge$50\% of that Gaussian's votes ($\ge$1 vote). If fewer than 5\% of
Gaussians end up labeled, the tolerance is tripled once and then the
visibility test is dropped, so a degenerate depth render cannot starve
the estimator.

\paragraph{Instances and extents.} A class is measured only if
$\ge$50 Gaussians carry its label. Instances come from voxel-grid
connected components (26-neighborhood) at voxel size $4\times$ the median
nearest-neighbor distance of that class's points (subsampled to 1500
points with a fixed seed), keeping at most the 4 largest; instances below
$0.5\times$ the largest instance's extent are discarded as fragments.
Extents are $[0.02,0.98]$ quantile-trimmed spans along the vertical axis.
A whole-cluster extent is added only when the class has at most one
instance, with half weight; each surviving instance also carries half
weight, split equally among instances of that class.

\paragraph{Vertical axis and planes.} Candidates are the floor-plane
normal, the ceiling-plane normal (each from a PCA on points within the
90th percentile of distance from the class median, requiring $\ge$30
points) and the global PCA minor axis; we keep whichever maximizes the
floor-to-ceiling separation. Plane heights are the modes of 60-bin
height histograms. A floor$\to$ceiling gap is rejected if it is below
15\% of the 5--95 percentile span of labeled heights, which removes
degenerate gaps from mislabeled ceilings.

\paragraph{Fusion.} Geman--McClure with $c{=}3$ on
$(e^{s}m_c-\mu_c)/\sigma_c$, plus the quadratic warm prior
($\sigma_w{=}0.35$, $\lambda{=}1$); the post-hoc estimate is the minimum
over 2001 log-spaced points in $e^{s}\in[0.05,20]$. In training the same
objective is minimized by 25 Adam steps at learning rate
$5\times10^{-3}$ every 200 iterations, over measurements buffered from
the last $K{=}5$ votes. Abstention triggers at inlier mass below
$M_{\min}{=}1.5$.

\begin{table}[htbp]
  \centering\scriptsize
  \caption{Complete anchor catalog (meters). \emph{what} is the
  measured quantity: \textsc{v} = vertical extent of the instance,
  \textsc{g} = gap between two plane modes. Every extent is the
  $[0.02,0.98]$ quantile-trimmed span along the semantic vertical axis;
  instance cues come from voxel connected components, cluster cues from
  the whole labeled set and only when the class has at most one
  instance. Weights encode how standardized a class is and were set once
  from public building standards and furniture dimensions.}
  \label{tab:supp-priors}
  \begin{tabular}{lcccc}
    \toprule
    class & what & $\mu$ & $\sigma$ & $w$ \\
    \midrule
bed & \textsc{v} &  0.55 & 0.15 & 0.30 \\
chair & \textsc{v} &  0.88 & 0.15 & 0.40 \\
coffee table & \textsc{v} &  0.45 & 0.10 & 0.30 \\
counter & \textsc{v} &  0.90 & 0.10 & 0.50 \\
countertop & \textsc{v} &  0.90 & 0.08 & 0.50 \\
desk & \textsc{v} &  0.75 & 0.08 & 0.60 \\
door & \textsc{v} &  2.03 & 0.10 & 1.00 \\
monitor & \textsc{v} &  0.45 & 0.15 & 0.20 \\
person & \textsc{v} &  1.70 & 0.15 & 0.30 \\
radiator & \textsc{v} &  0.60 & 0.20 & 0.20 \\
refrigerator & \textsc{v} &  1.70 & 0.25 & 0.40 \\
sink & \textsc{v} &  0.85 & 0.12 & 0.40 \\
stool & \textsc{v} &  0.65 & 0.15 & 0.30 \\
swivel chair & \textsc{v} &  1.00 & 0.20 & 0.30 \\
table & \textsc{v} &  0.74 & 0.08 & 0.60 \\
television & \textsc{v} &  0.70 & 0.30 & 0.15 \\
toilet & \textsc{v} &  0.76 & 0.10 & 0.50 \\
wardrobe & \textsc{v} &  2.00 & 0.30 & 0.30 \\
floor$\to$ceiling gap & \textsc{g} &  2.50 & 0.30 & 1.00 \\
    \bottomrule
  \end{tabular}
\end{table}

\section{Qualitative results and diagnostics}
This section collects the visual evidence behind the main paper's claims:
that the scale machinery is quality-neutral on the scenes it trains
(render grid), that the in-loop estimate converges early and stays put
(trajectories), that the real-SfM condition reproduces the sim3 pattern
scene-by-scene (scatter), what the forced-scale degradation
(Tab.~\ref{tab:supp-e2}; Sec.~4.5 of the main paper) looks like at
full resolution (frame pair), and what the teacher actually
sees in each abstention regime (gallery). The last is the
qualitative counterpart of the two-level coverage evidence in Sec.~4.6
of the main paper.
\begin{figure}[htbp]
  \centering
  \includegraphics[width=\linewidth]{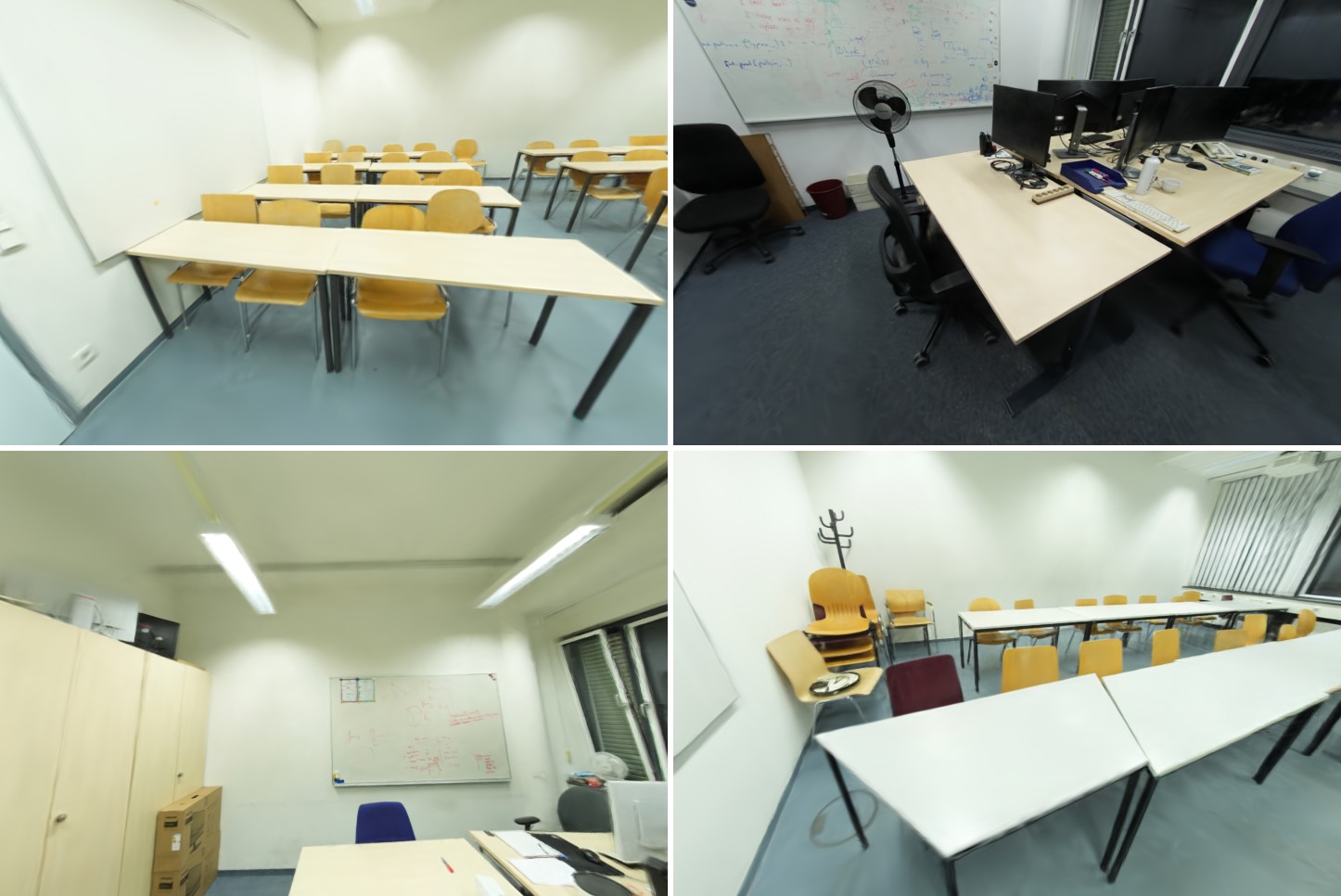}
  \caption{Renders of four ScanNet++ scenes trained with the full
  scale-in-the-loop pipeline; the scale machinery costs no visible quality
  (mean $\Delta$PSNR within $0.07$\,dB of the ablated baselines;
  Tab.~\ref{tab:supp-quality}).}
\end{figure}
\begin{figure}[htbp]
  \centering
  \includegraphics[width=0.95\linewidth]{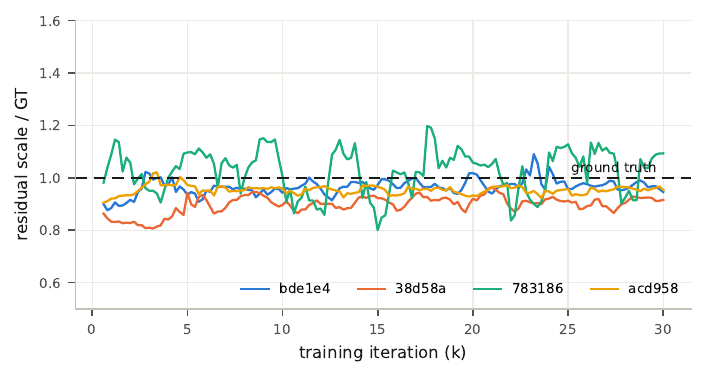}
  \caption{In-loop residual-scale trajectories (normalized by ground
  truth) for four scenes: the estimate stabilizes within the first third of
  training; the reported value is the post-hoc multi-seed MLE.}
\end{figure}
\begin{figure}[htbp]
  \centering
  \includegraphics[width=0.7\linewidth]{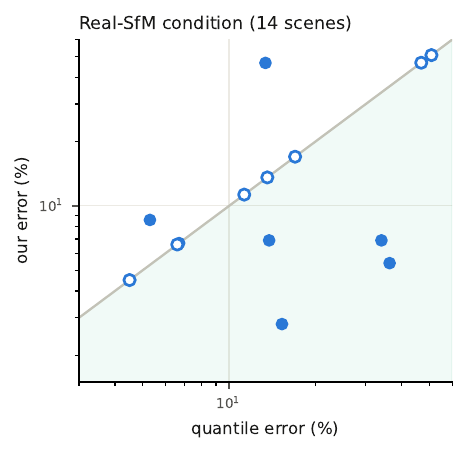}
  \caption{Real-SfM condition, per scene (open circles = abstentions).}
\end{figure}
\begin{table}[htbp]
  \centering\scriptsize
  \caption{Forced-scale training (fixed hyperparameters, 3 development
  scenes; Sec.~4.5 of the main paper). Gross under-scaling is uniformly
  harmful in quality and destabilizes the Gaussian budget in a
  scene-dependent direction; differences within the
  nominal-to-$10\times$ range are at seed-noise level.}
  \label{tab:supp-e2}
  \begin{tabular}{lcccc}
    \toprule
    & $0.1\times$ & nominal & $\approx$GT & $10\times$ \\
    \midrule
    \multicolumn{5}{l}{\emph{PSNR}$\uparrow$} \\
    room1   & 26.73 & 27.39 & \textbf{27.67} & 27.61 \\
    room2   & 28.86 & 29.89 & \textbf{29.97} & 29.58 \\
    studio1 & 22.56 & \textbf{26.70} & 26.33 & 26.45 \\
    \midrule
    \multicolumn{5}{l}{\emph{Gaussians (k)}$\downarrow$} \\
    room1   & 376 & 299 & \textbf{286} & 287 \\
    room2   & 442 & 316 & \textbf{316} & 320 \\
    studio1 & \textbf{251} & 370 & 373 & 356 \\
    \bottomrule
  \end{tabular}
\end{table}
\begin{figure}[htbp]
  \centering
  \includegraphics[width=0.48\linewidth]{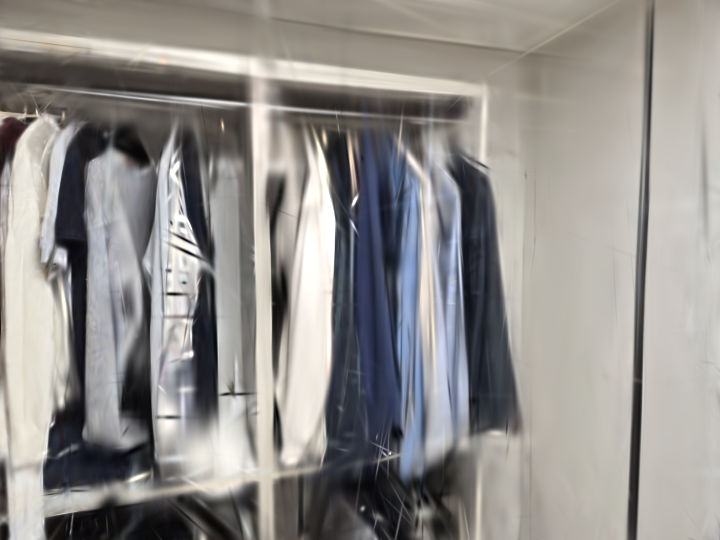}\hfill
  \includegraphics[width=0.48\linewidth]{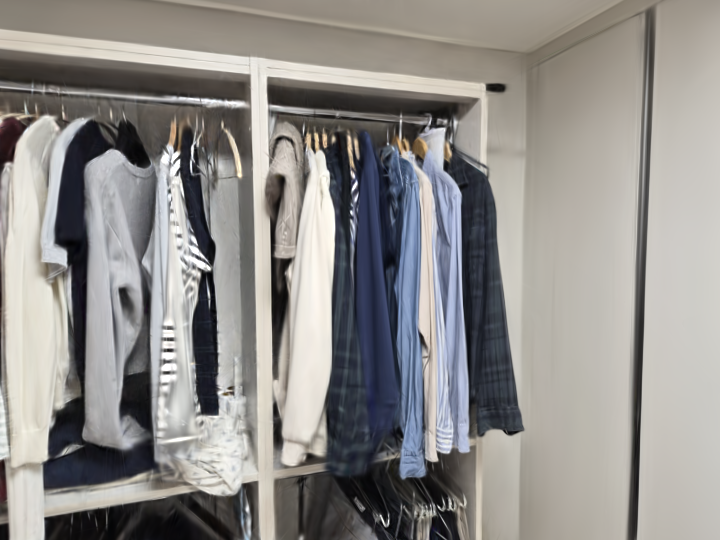}
  \caption{Full frames behind the forced-scale experiment
  (Tab.~\ref{tab:supp-e2}; Sec.~4.5 of the main paper): the same view
  trained at $0.1\times$ the metric scale (left) and at the metric
  scale (right).}
\end{figure}
\begin{figure}[htbp]
  \centering
  \includegraphics[width=\linewidth]{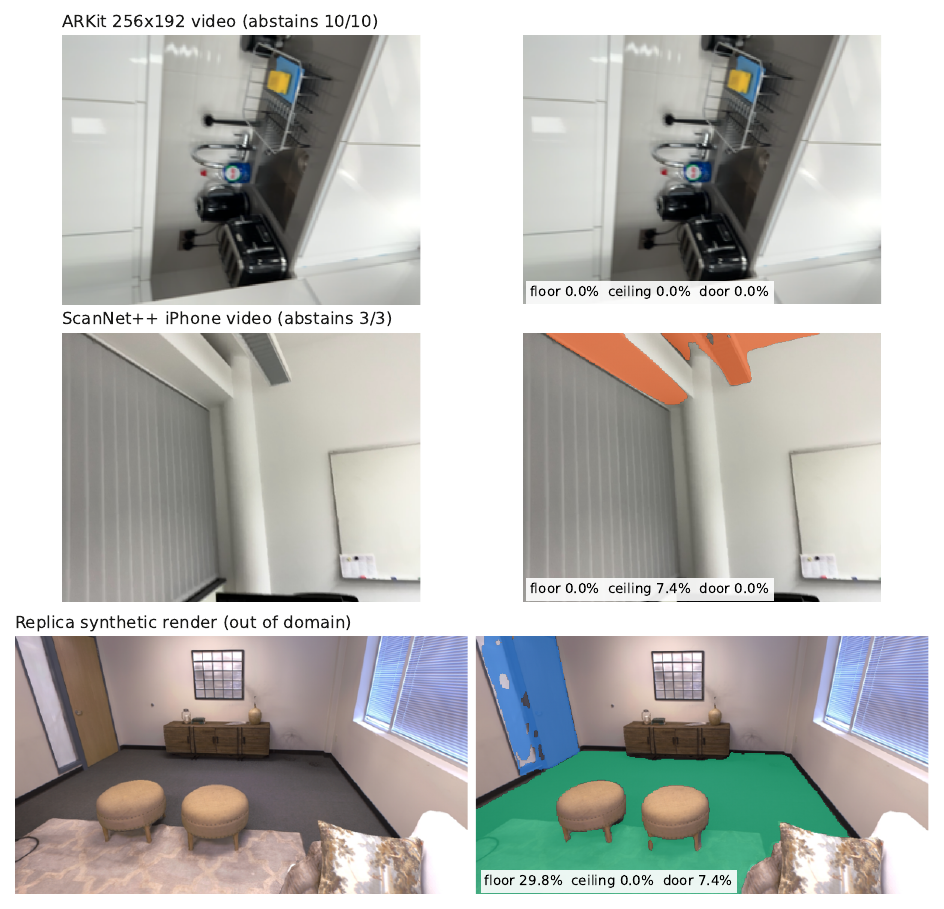}
  \caption{Abstention regimes with the teacher's own output. Top:
  ARKitScenes $256{\times}192$ video, where anchors are labeled and survive
  into 3D, but the resulting measurements disagree
  (Sec.~\ref{supp:ladder}). Middle: ScanNet++ iPhone video with motion
  blur. Bottom: a Replica synthetic render, where the real-image teacher
  fails outright. In all three the estimator abstains.}
\end{figure}

\section{Mechanism visualizations}
The following figures visualize the individual mechanisms in place: the
3D anchors and the vertical axis they define, the one-sided visibility
test that keeps grazing-incidence floor votes alive, the abstention gate
over the empirical inlier-mass distribution (shown \emph{with} the
degraded actives it fails to separate; consistent with the failure
analysis, their cues agree while being collectively biased), the
two-level coverage evidence, and the cost--accuracy landscape the router
operates in (the nominal-height sensitivity curve is shown with its
retraining study, Fig.~\ref{fig:supp-height}).
\begin{figure}[htbp]
  \centering
  \includegraphics[width=\linewidth]{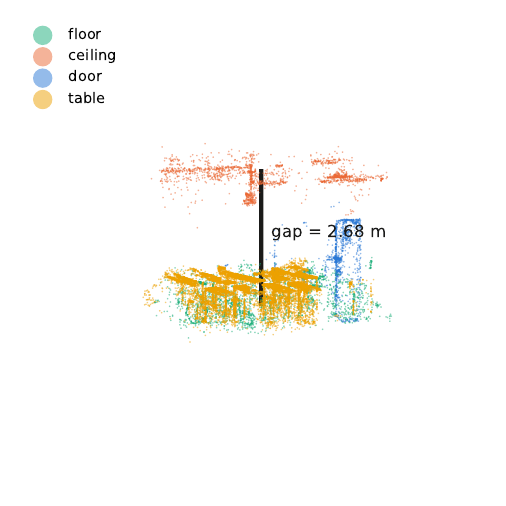}
  \caption{Labeled Gaussians of a ScanNet++ scene in 3D (subsampled),
  with the semantically derived vertical axis and the measured
  floor$\to$ceiling gap under the recovered scale.}
\end{figure}
\begin{figure}[htbp]
  \centering
  \includegraphics[width=\linewidth]{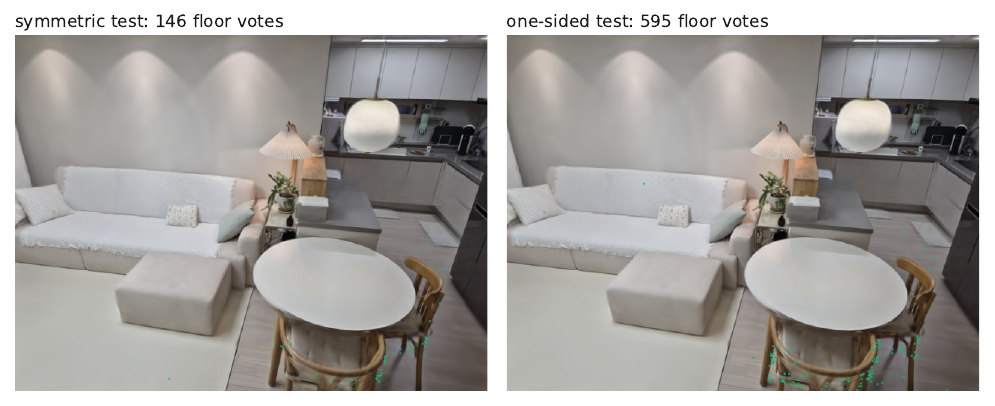}
  \caption{Why the visibility test must be one-sided: on the same model
  and identical votes, the symmetric $|z-D|$ test keeps only 146
  floor-labeled Gaussians (grazing-incidence rejections), the one-sided
  test 595: the difference between a collapsed and a usable gap cue.}
\end{figure}
\begin{figure}[htbp]
  \centering
  \includegraphics[width=0.9\linewidth]{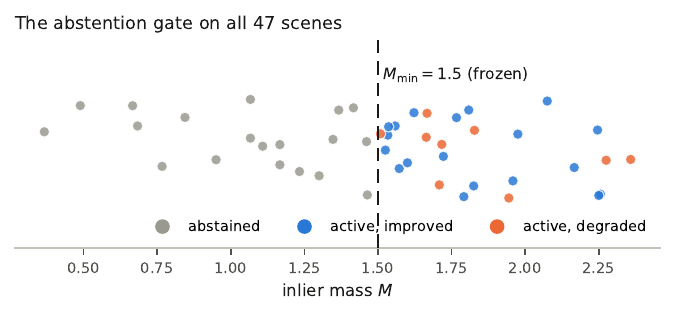}
  \caption{Inlier mass of every scene against the frozen threshold.
  Degraded actives (orange) sit \emph{inside} the accepted region ---
  the gate cannot separate them, matching the failure analysis: their
  cues agree with each other while being collectively biased.}
\end{figure}
\begin{figure}[htbp]
  \centering
  \includegraphics[width=0.9\linewidth]{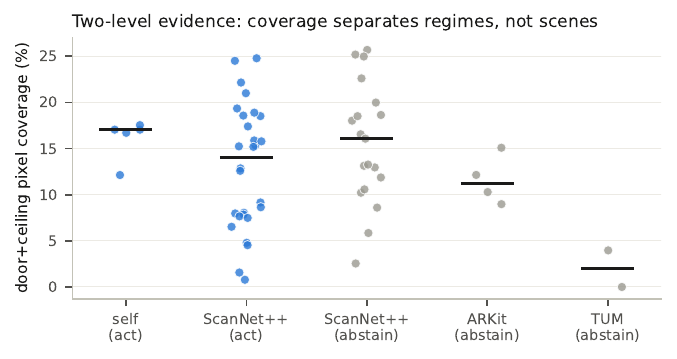}
  \caption{Two-level abstention evidence: the teacher's 2D anchor
  coverage separates capture regimes (right three columns) but not
  individual ScanNet++ scenes.}
\end{figure}
\begin{figure}[htbp]
  \centering
  \includegraphics[width=0.55\linewidth]{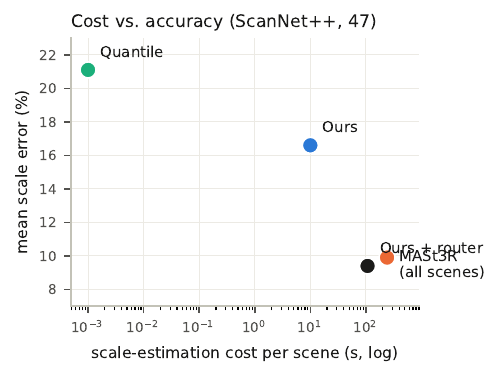}
  \caption{Cost--accuracy trade-off among the estimators that run inside
  the reconstruction pipeline; the router occupies the corner. Single-image
  metric models are not plotted: on ScanNet++ MoGe-2 would sit below and
  left of every point here, but that benchmark is inside its training
  mixture (Sec.~\ref{supp:indomain}), so the position would not be
  comparable.
  The $x$-axis is scale-estimation cost per scene and excludes 3DGS
  training, which the deployment scenario of the main paper (Sec.~4.1)
  treats as sunk: the reconstruction is built regardless.}
\end{figure}

\section{Cue composition and prior sensitivity}
Across the 28 active ScanNet++ scenes, the floor$\to$ceiling gap carries
$\sim$54\% of the inlier mass regardless of outcome; the discriminative
difference is doors (15\% of inlier mass on improved scenes vs.\ 6\% on
degraded ones). Table~\ref{tab:supp-cuecomp} gives the cue-composition
ablation referenced in the main paper: single-cue estimators match the
full system's median while their mean collapses, because neither has any
defense on the scenes where its one cue fails.

\begin{table}[htbp]
  \centering\scriptsize
  \caption{Cue-composition ablation, ScanNet++ 47 (deployed scale
  error). Single-cue variants track the median but not the mean.}
  \label{tab:supp-cuecomp}
  \begin{tabular}{lcc}
    \toprule
    estimator & mean & median \\
    \midrule
    room-height prior only    & 40.5\% & 10.7\% \\
    gap cue only              & 39.6\% & 9.4\% \\
    uniform cue weights       & 47.9\% & 13.1\% \\
    \textbf{full fusion}      & \textbf{16.6\%} & \textbf{10.6\%} \\
    \bottomrule
  \end{tabular}
\end{table}

\paragraph{Leave-one-anchor-out.}
Table~\ref{tab:supp-loo} removes one anchor class at a time from the
frozen cue sets and re-runs only the MLE and gate. No single furniture
anchor is load-bearing: the largest movement is doors at $+1.8$\,pp, and
dropping \texttt{person} reproduces the $16.6{\to}16.8\%$ figure of
Sec.~\ref{supp:custom}. Removing the floor$\to$ceiling gap does not
degrade the estimate; it removes the estimate. All 47 scenes abstain and
inherit the warm start (21.1\%), which is the gate behaving as designed:
the failure mode of losing the load-bearing cue is refusal, not a
confident wrong answer. The complementary question --- whether the gap
alone would suffice --- is answered above (gap-only, 39.6\% mean against
16.6\% for the full fusion).

\begin{table}[htbp]
  \centering\scriptsize
  \caption{Leave-one-anchor-out on the frozen ScanNet++ cue sets (47
  scenes, deployed protocol: abstentions inherit the warm estimate).
  Each row removes every measurement of one anchor class and re-runs
  only the MLE and gate.}
  \label{tab:supp-loo}
  \setlength{\tabcolsep}{2.6pt}
  \begin{tabular}{lccc@{\hskip 8pt}lccc}
    \toprule
    removed & mean & med. & abst. & removed & mean & med. & abst. \\
    \midrule
    none (frozen) & 16.6 & 10.6 & 19 & person & 16.8 & 10.6 & 18 \\
    floor$\to$ceil.\ gap & 21.1 & 17.5 & 47 & stool & 16.7 & 10.6 & 20 \\
    door & 18.5 & 13.8 & 25 & refrigerator & 16.7 & 10.6 & 19 \\
    chair & 18.1 & 12.0 & 22 & swivel chair & 16.7 & 10.6 & 22 \\
    bed & 17.7 & 11.5 & 20 & toilet & 16.7 & 11.1 & 19 \\
    table & 17.6 & 12.4 & 22 & counter & 16.6 & 10.6 & 19 \\
    television & 17.5 & 11.5 & 22 & radiator & 16.6 & 10.6 & 21 \\
    desk & 17.3 & 11.6 & 25 & sink & 16.4 & 10.3 & 17 \\
    wardrobe & 17.1 & 11.1 & 19 & countertop & 16.4 & 9.8 & 18 \\
    monitor & 17.1 & 11.1 & 21 & coffee table & 17.0 & 11.1 & 20 \\
    \bottomrule
  \end{tabular}
\end{table}

\begin{figure}[htbp]
  \centering
  \includegraphics[width=\linewidth]{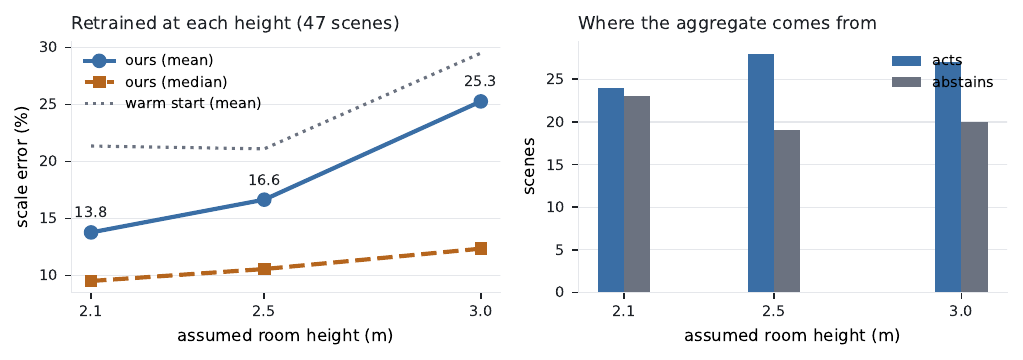}
  \caption{Measured sensitivity to the assumed room height (47 scenes
  retrained per setting). Left: the estimator tracks its warm start and
  beats it everywhere. Right: the active/abstained split, which is what
  moves the aggregates.}
  \label{fig:supp-height}
\end{figure}

\paragraph{Nominal room height, measured.}
The nominal height is an input to the quantile warm start, which rescales
the source before training, so its effect can only be measured by
retraining. We retrained all 47 scenes at 2.1\,m and 3.0\,m and evaluated
them with the unchanged estimator (Tab.~\ref{tab:supp-height},
Fig.~\ref{fig:supp-height}). Three things follow. First, most of the
effect is inherited rather than introduced: the heuristic itself moves
from 21.3\% to 29.5\% across the range, and our error tracks it.
Second, the improvement over the warm start survives every setting, from
$-36\%$ relative at 2.1\,m to $-14\%$ at 3.0\,m. Third, contrary
to the concern that a 2.5\,m prior is quietly matched to this dataset,
2.5\,m is \emph{not} the best-performing choice here: 2.1\,m would have
been better on all three aggregates. We kept 2.5\,m because it is the
architectural standard we froze before looking, and report that a
tuned choice would have flattered us.

\begin{table}[htbp]
  \centering\scriptsize
  \caption{Assumed room height, all 47 scenes \emph{retrained} at each
  setting. ``active'' is the mean over scenes the estimator acts on.}
  \label{tab:supp-height}
  \begin{tabular}{lccccc}
    \toprule
    height & quantile & ours & median & active & abstain \\
    \midrule
    2.1\,m & 21.3\% & \textbf{13.8\%} & \textbf{9.5\%} & \textbf{7.4\%} & 23 \\
    2.5\,m (frozen) & 21.1\% & 16.6\% & 10.6\% & 9.3\% & 19 \\
    3.0\,m & 29.5\% & 25.3\% & 12.3\% & 13.1\% & 20 \\
    \bottomrule
  \end{tabular}
\end{table}

\paragraph{Adapting the priors to another region.}
The catalog encodes one building tradition, and a practitioner
elsewhere should expect to edit it. Two measurements bound what that
costs. Shifting \emph{every} prior mean by $\pm10\%$, a larger move
than the difference between most national door standards, changes the
47-scene mean only from $16.6\%$ to $17.9$--$18.5\%$; the robust
fusion absorbs a uniform prior shift because the cues outvote it.
Discarding the reliability weights, by contrast, costs $31$\,pp. Our
recommendation is therefore to keep the weighting tiers as they are and
edit only the two code-regulated means ($\mu$ for \texttt{door} and the
floor$\to$ceiling gap) to local standards; the ergonomic and
style-dependent classes carry little mass and are not worth
regionalising. The one input that does warrant local calibration is the
nominal room height of the warm start, whose effect is measured above.

Changing the size prior's \emph{center} rather than the warm start is a
separate, milder knob. Refitting the post-hoc estimator with the
floor-to-ceiling prior re-centered (warm start untouched) moves the
47-scene mean from 16.6\% to 19.2\% at 2.1\,m and 20.5\% at 3.0\,m; the
result is likewise insensitive to that prior's width.

\section{A tried-and-rejected variant: asymmetric robust loss}
\label{supp:asym}
Because instance extents are lower bounds and cluster extents upper
bounds, an asymmetric Geman--McClure penalty (early saturation on the
expected side of each cue type) appears principled. We implemented it and
it \emph{regressed both} our development scenes and the public benchmarks:
the asymmetry double-counts the bias cancellation that the hybrid
cluster+instance cue set already provides, and weakens the anchors exactly
where mode selection needs them. We report this as a negative design
result; learned per-cue reliabilities remain future work.

\section{Additional robustness data}
Tables~\ref{tab:supp-tau}--\ref{tab:supp-perturb} give the full
abstention-threshold sweep and the prior-perturbation results on all 47
scenes. Seed-ensemble repeats vary the estimate by $\pm0.13$\,pp over
three evaluation repeats; training re-runs vary active corrections by
$<1.5$\,pp and occasionally convert a borderline scene to abstention.
Per-scene cue-coverage statistics ship with the code release.

\begin{table}[htbp]
  \centering\scriptsize
  \caption{Abstention-threshold sweep, ScanNet++ 47 (deployed protocol:
  abstentions inherit the warm estimate). The frozen $M_{\min}{=}1.5$
  sits on the mean plateau; pushing the threshold up converts the
  benchmark into the warm baseline.}
  \label{tab:supp-tau}
  \begin{tabular}{lccc}
    \toprule
    $M_{\min}$ & mean & median & \#abstain \\
    \midrule
    0.50 & 17.1\% & 8.7\% & 2 \\
    0.75 & 17.3\% & 8.7\% & 4 \\
    1.00 & 15.5\% & 8.7\% & 7 \\
    1.25 & 15.2\% & 9.8\% & 13 \\
    \textbf{1.50 (frozen)} & \textbf{16.6\%} & \textbf{10.6\%} & \textbf{19} \\
    1.75 & 17.9\% & 12.9\% & 32 \\
    2.00 & 19.2\% & 15.0\% & 40 \\
    2.50 & 21.1\% & 17.5\% & 47 \\
    \bottomrule
  \end{tabular}
\end{table}

\begin{table}[htbp]
  \centering\scriptsize
  \caption{Prior perturbations, ScanNet++ 47.}
  \label{tab:supp-perturb}
  \begin{tabular}{lccc}
    \toprule
    variant & mean & median & \#abstain \\
    \midrule
    frozen configuration & 16.6\% & 10.6\% & 19 \\
    all prior means $+10\%$ & 17.9\% & 13.7\% & 22 \\
    all prior means $-10\%$ & 18.5\% & 13.9\% & 17 \\
    gap prior $\sigma{=}0.45$ & 17.3\% & 9.8\% & 16 \\
    gap prior $\sigma{=}0.60$ & 16.9\% & 10.2\% & 17 \\
    gap prior $\sigma{=}0.90$ & 17.0\% & 11.1\% & 16 \\
    uniform cue weights & 47.9\% & 13.1\% & 0 \\
    \bottomrule
  \end{tabular}
\end{table}

\paragraph{Heterogeneous prior perturbation.}
\label{supp:locale}
The perturbation rows of Tab.~\ref{tab:supp-perturb} shift every anchor mean
by the same factor. That is the easy case: a common multiplicative shift on
every $\mu$ is exactly a shift of the global scale the estimator is solving
for, so the fusion only has to move its single free variable. A practitioner
working from a different regional catalog faces something harder --- door
leaves at 1.80--2.00\,m against the 2.03\,m we assume, ceilings at 2.40\,m
against 2.50\,m, table heights barely moving --- that is, cue-specific shifts
of differing sign which cannot be absorbed by one scalar and instead appear
as genuine disagreement between cues.

Tab.~\ref{tab:supp-locale} measures that. Only the prior means are changed;
sigmas and reliability weights are untouched, and the estimator, the gate and
the trained models are the frozen ones. Everything is recomputed from the
saved cue sets, so the frozen and uniform rows reproduce
Tab.~\ref{tab:supp-perturb} exactly and act as a check on the recomputation.

Two named catalogs (EU and JP conventions) and a signed adversarial example
all cost \emph{less} than the uniform shifts. Drawing every anchor mean
independently from $U[\pm10\%]$, 200 times, gives a median mean-error of 17.1\%
(90\% interval [16.2, 17.9], worst draw 18.7\%), and none of the 200 draws
is worse than the quantile heuristic it would replace. Widening to $U[\pm20\%]$ gives
median 18.1\% and worst 21.7\%, with 2\% of draws worse than the heuristic.
Independent shifts partly cancel where a common one does not, and the robust
loss downweights the disagreeing minority.

\begin{table}[htbp]
  \centering\scriptsize
  \caption{Heterogeneous prior perturbation, ScanNet++ 47.}
  \label{tab:supp-locale}
  \begin{tabular}{lccc}
    \toprule
    catalog & mean & median & \#abstain \\
    \midrule
    frozen configuration & 16.6\% & 10.6\% & 19 \\
    all means $+10\%$ (uniform) & 17.9\% & 13.7\% & 22 \\
    all means $-10\%$ (uniform) & 18.5\% & 13.9\% & 17 \\
    \midrule
    EU catalog (heterogeneous) & 16.9\% & 11.1\% & 19 \\
    JP catalog (heterogeneous) & 17.3\% & 11.9\% & 18 \\
    signed adversarial example & 17.2\% & 12.2\% & 17 \\
    \bottomrule
  \end{tabular}
\end{table}

\section{Resolution ladder on held-out video}
\label{supp:ladder}
Table~\ref{tab:supp-ladder} reports the paired resolution experiment
behind the refuted explanation in the main paper. The two conditions
share each video's trajectory, depth-seeded points and sim(3) seed, so
only the image resolution differs; the warm start differs slightly
because it is computed on the resulting reconstruction. Cue mass rises
in six of ten videos and reaches 1.43 in one, but no video crosses the
frozen 1.5 gate, and the median cue residual is essentially unchanged
($5.31\sigma \to 5.55\sigma$). We report this as a negative result: it
refutes our resolution hypothesis and relocates the cause to cue
disagreement under fragmented, handheld-video reconstruction.

\paragraph{Scaling the held-out study to 50 videos.}
After the ten pre-registered videos, we extended the study by forty more
ARKitScenes Validation videos selected blind --- SHA-1 of the video id,
ascending, over the videos not previously downloaded; no attribute
filtering, no content inspection --- with the frozen pipeline verbatim
and one evaluation per video, after logging the same prediction
(overwhelming abstention, zero degradation). The outcome: 40/40 abstain,
the delivered estimate equals the warm start on every video, and no
video degrades. Cues are present (median 32 per video, none cue-less);
the inlier mass simply never approaches the gate (mean 0.67, maximum
1.24 against the frozen 1.5). Across all fifty held-out videos the
abstention rate is 50/50 and the inherited warm error averages 14.1\%.

\begin{table}[htbp]
  \centering\scriptsize
  \caption{Held-out ARKitScenes at two resolutions (frozen
  configuration). Every video abstains in both conditions, so the
  delivered error equals the warm start's.}
  \label{tab:supp-ladder}
  \begin{tabular}{lcccccc}
    \toprule
    & \multicolumn{3}{c}{$256{\times}192$} & \multicolumn{3}{c}{$640{\times}480$} \\
    \cmidrule(lr){2-4}\cmidrule(lr){5-7}
    video & mass & cues & err & mass & cues & err \\
    \midrule
\texttt{41125709} & 0.53 & 27 & 15.5 & 0.53 & 27 & 7.4 \\
\texttt{41125718} & 0.36 & 23 & 6.8 & 0.22 & 28 & 7.3 \\
\texttt{41125722} & 0.16 & 33 & 6.9 & 0.24 & 28 & 5.4 \\
\texttt{41125731} & 0.38 & 26 & 10.5 & 0.05 & 19 & 11.9 \\
\texttt{41125756} & 0.28 & 25 & 0.5 & 0.68 & 33 & 0.5 \\
\texttt{41125760} & 0.38 & 35 & 3.4 & 1.43 & 34 & 13.2 \\
\texttt{41125763} & 0.27 & 39 & 11.5 & 0.30 & 43 & 6.4 \\
\texttt{41142278} & 0.40 & 35 & 33.1 & 0.50 & 24 & 52.1 \\
\texttt{41142280} & 0.35 & 31 & 0.7 & 0.00 & 21 & 4.5 \\
\texttt{41142281} & 0.12 & 16 & 5.2 & 0.17 & 16 & 5.7 \\
    \midrule
    mean & 0.33 & 29 & 9.4 & 0.41 & 27 & 11.4 \\
    \bottomrule
  \end{tabular}
\end{table}

\section{Training-component ablation}
\label{supp:ablation19}
Table~\ref{tab:supp-abl19-agg} gives the aggregate training-loop
component ablation referenced in the main paper, with the paired
bootstrap CIs behind the reported null: no component changes deployed
scale error significantly on the 19-scene development split (the
ScanNet++ scenes evaluated first; the configuration itself was frozen on
the five self-captured scenes, main paper Sec.~4.1).
Table~\ref{tab:supp-abl19} gives the per-scene numbers. They show
directly why we withdrew the ordering our earlier 3-scene version
suggested: the largest per-scene gaps are abstention flips
(\texttt{fb5a96b1a2}: the full system abstains at 24.9\% where the
no-in-loop variant acts at 2.3\%; \texttt{1ada7a0617}: the reverse), not
consistent accuracy differences. The one qualitative asymmetry is
\texttt{f9f95681fd}, where removing the depth regularizer produces a
confidently wrong 47.0\% active correction on a scene every other
variant abstains from.

\begin{table}[htbp]
  \centering\scriptsize
  \caption{Training-loop component ablation, aggregate (ScanNet++
  19-scene development split, deployed scale error). Paired bootstrap
  95\% CIs of the difference vs.\ full; none excludes zero
  (all $p{\ge}0.23$).}
  \label{tab:supp-abl19-agg}
  \begin{tabular}{lccc}
    \toprule
    variant & mean & median & CI vs.\ full (pp) \\
    \midrule
    full              & 11.0\% & 7.0\% & --- \\
    no in-loop scale  & 9.8\%  & 5.5\% & $[-4.1,+1.1]$ \\
    no depth reg.     & 12.7\% & 9.7\% & $[-2.5,+6.3]$ \\
    no metric densif. & 11.2\% & 8.1\% & $[-0.3,+0.8]$ \\
    \bottomrule
  \end{tabular}
\end{table}

\begin{table}[htbp]
  \centering\scriptsize
  \caption{Per-scene deployed scale error (\%), 19-scene development
  split. $\dagger$ marks abstention (estimate inherited from the warm
  start).}
  \label{tab:supp-abl19}
  \begin{tabular}{lcccc}
    \toprule
    scene & full & no in-loop & no depth reg. & no metric dens. \\
    \midrule
\texttt{1ada7a0617} & 5.3 & 15.0$^\dagger$ & 15.0$^\dagger$ & 5.6 \\
\texttt{21d970d8de} & 9.8 & 7.8 & 5.3 & 10.0 \\
\texttt{38d58a7a31} & 10.6 & 8.4 & 11.6 & 10.3 \\
\texttt{3e8bba0176} & 7.0 & 9.2 & 11.4 & 8.1 \\
\texttt{3f15a9266d} & 2.1 & 2.4 & 1.8 & 1.8 \\
\texttt{40aec5fffa} & 47.2$^\dagger$ & 47.2$^\dagger$ & 47.2$^\dagger$ & 47.2$^\dagger$ \\
\texttt{5748ce6f01} & 4.1 & 2.3 & 4.4 & 3.8 \\
\texttt{5eb31827b7} & 8.7 & 4.5$^\dagger$ & 12.2 & 8.4 \\
\texttt{5ee7c22ba0} & 8.6 & 5.4 & 14.2 & 6.9 \\
\texttt{7831862f02} & 4.7 & 5.2 & 0.2 & 4.3 \\
\texttt{7b6477cb95} & 5.1 & 4.8 & 10.4 & 7.6 \\
\texttt{a24f64f7fb} & 5.5 & 5.5 & 3.9 & 3.3 \\
\texttt{acd95847c5} & 4.1 & 3.5 & 3.8 & 4.1 \\
\texttt{bde1e479ad} & 5.7 & 5.7 & 7.7 & 8.5 \\
\texttt{c4c04e6d6c} & 27.8$^\dagger$ & 27.8$^\dagger$ & 27.8$^\dagger$ & 27.8$^\dagger$ \\
\texttt{c50d2d1d42} & 8.7 & 10.8 & 9.7 & 11.6 \\
\texttt{cc5237fd77} & 4.9 & 4.6 & 6.1 & 4.6 \\
\texttt{f9f95681fd} & 14.0$^\dagger$ & 14.0$^\dagger$ & 47.0 & 14.0$^\dagger$ \\
\texttt{fb5a96b1a2} & 24.9$^\dagger$ & 2.3 & 2.0 & 24.9$^\dagger$ \\
    \midrule
    mean / median & 11.0 / 7.0 & 9.8 / 5.5 & 12.7 / 9.7 & 11.2 / 8.1 \\
    \bottomrule
  \end{tabular}
\end{table}

\section{Does the abstention gate need recalibrating per teacher?}
\label{supp:gate-teacher}
The gate is a fixed threshold on a quantity (inlier cue mass) that a
better segmenter should inflate, so it is fair to ask whether
$M_{\min}{=}1.5$ is specific to the R50 teacher and whether a
teacher-agnostic calibration is needed. We swept the threshold under
both teachers on the same 47 trained models, recording the estimate
\emph{before} the gate is applied so the sweep can run in both
directions (Tab.~\ref{tab:supp-gate-teacher}).

\begin{table}[htbp]
  \centering\scriptsize
  \caption{Abstention threshold under two teachers (ScanNet++ 47,
  deployed protocol: abstentions inherit the warm estimate). The optimum
  does not move with teacher strength.}
  \label{tab:supp-gate-teacher}
  \begin{tabular}{lcccc}
    \toprule
    & \multicolumn{2}{c}{R50 (47.2 mIoU)} & \multicolumn{2}{c}{Swin-L (57.3 mIoU)} \\
    \cmidrule(lr){2-3}\cmidrule(lr){4-5}
    $M_{\min}$ & mean & \#abstain & mean & \#abstain \\
    \midrule
    0.50 & 17.1\% & 2 & 18.5\% & 1 \\
    0.75 & 17.3\% & 4 & 17.8\% & 3 \\
    1.00 & 15.5\% & 7 & 16.7\% & 4 \\
    1.25 & \textbf{15.2\%} & 13 & \textbf{15.2\%} & 14 \\
    \textbf{1.50 (frozen)} & 16.6\% & 19 & 15.7\% & 17 \\
    1.75 & 17.9\% & 32 & 17.6\% & 30 \\
    2.00 & 19.2\% & 40 & 19.1\% & 39 \\
    2.50 & 21.1\% & 47 & 21.1\% & 47 \\
    \bottomrule
  \end{tabular}
\end{table}

The answer is that no per-teacher calibration is required. Both curves
are shallow basins with the same minimum at $M_{\min}{=}1.25$ and the
same value there ($15.2\%$), and both degrade in the same way above
$1.5$. Swapping in a segmenter ten mIoU stronger shifts the individual
scene masses but not the operating point, so a temperature scaling on
inlier mass would be correcting a drift that does not occur.

What does change is the price of our conservatism. The frozen $1.5$
costs $1.4$\,pp against the best threshold under R50 and only
$0.5$\,pp under Swin-L: a better teacher produces cue sets that agree
more, so a slightly-too-strict gate throws away less. We report this
rather than retuning, and it is the second place in this paper where
freezing before evaluation cost us accuracy.

One caveat on how this table was produced. Sweeping \emph{below} the
frozen threshold requires the estimate the fusion would have delivered
on scenes it actually refused, which the deployed code discards. We
added a read-only field recording the pre-gate estimate; no decision
reads it and the deployed numbers elsewhere in this paper are
unaffected. An earlier version of this analysis mistakenly swept the
post-gate estimate, which reproduces the warm baseline on every refused
scene and makes the curve look flat below $1.5$; the table above is the
corrected version.

\section{Does softer visibility reasoning help?}
\label{supp:softvis}
Occlusion-aware label lifting for 3DGS has been treated far more
carefully elsewhere: LabelGS runs an explicit occlusion analysis, and
VALA weights each ray by a Gaussian's marginal contribution. Our gate is
a single hard depth comparison, which invites the question of whether a
softer, contribution-style test would recover more usable cues. We
tested the closest variant we can implement without modifying the
rasterizer: instead of admitting a vote only when
$z < D_v(1+\tau_{\mathrm{vis}})$, every projection votes with a
transmittance-style weight
\begin{equation}
  w = \exp\!\Big(\!-\tfrac{1}{2}
      \big[\tfrac{\max(0,\,z - D_v)}{\tau_{\mathrm{vis}} D_v}\big]^2\Big),
  \label{eq:softvis}
\end{equation}
so a Gaussian at or in front of the rendered surface votes fully and one
behind it votes with exponentially decaying weight rather than not at
all. We stress that this is our own soft gate and not a reimplementation
of either published method: both operate on language features and VALA's
marginal contribution requires per-ray transmittance that the stock
rasterizer does not expose.

Softening does not help (Tab.~\ref{tab:supp-vis}). It is worse than the
hard one-sided test on every deployed number: mean $17.6\%$ against
$16.6\%$, median $11.6\%$ against $10.6\%$, three more abstentions, three
fewer actionable scenes, and a router that costs more calls for a worse
result ($9.8\%$ at 22 calls against $9.4\%$ at 19). It lands between the
one-sided and symmetric tests, which is what the mechanism predicts:
partial votes from points behind the surface are exactly the
occluder-contaminated votes the gate exists to remove, and readmitting
them at reduced weight readmits a fraction of the label spillover.

The reason a hard gate suits this task is that we measure \emph{extents},
not appearance. A soft weight is well matched to feature aggregation,
where a slightly wrong blend degrades gracefully. An extent is a
quantile of a point set, so a small number of occluder-labeled points at
an instance's extremes moves the measurement by far more than their
weight suggests. We therefore keep the hard test, and note that the
principled visibility machinery of LabelGS and VALA remains the natural
front-end for the partial-instance failure mode of the main paper ---
a direction we have not combined with ours.

\section{Reconstruction quality under metric densification}
\label{supp:quality}
Anchoring densification in meters changes what the reconstruction does
during training, so it is fair to ask what it costs. We compare the full
system against each training-component ablation on the 19-scene
development split, reading the final metrics of runs that already exist
(Tab.~\ref{tab:supp-quality}); no retraining was performed.

\begin{table}[htbp]
  \centering\scriptsize
  \caption{Rendering quality and Gaussian budget, full system vs.\ each
  training-component ablation (ScanNet++ 19, paired, final iteration).
  Means over scenes; $\Delta$ is full minus ablated.}
  \label{tab:supp-quality}
  \begin{tabular}{lccccc}
    \toprule
    & PSNR & SSIM & LPIPS & Gauss.\ (k) & $\Delta$PSNR \\
    \midrule
    full system         & 32.62 & 0.9508 & 0.1194 & 320 & --- \\
    no in-loop scale    & 32.55 & 0.9506 & 0.1197 & 322 & $+0.07$ \\
    no depth reg.       & 32.66 & 0.9510 & 0.1191 & 321 & $-0.04$ \\
    no metric densif.   & 32.68 & 0.9514 & 0.1185 & 322 & $-0.06$ \\
    \bottomrule
  \end{tabular}
\end{table}

Three points. The aggregate is neutral: every mean $\Delta$PSNR sits
within $0.07$\,dB, SSIM within $0.0006$ and LPIPS within $0.001$, so
metric densification neither buys nor costs rendering quality at the
scale these scenes are reconstructed at. The per-scene picture is
noisier than the aggregate suggests and we state it rather than the mean
alone: individual scenes move by up to $1.3$\,dB in either direction
(8 of 19 favor the full system against metric densification, 11 against
it), which is the ordinary seed-level variability of 3DGS training and
not a systematic effect. The Gaussian budget is likewise stable: the
median per-scene difference against the no-densification ablation is
$1.1\%$ (max $11.4\%$), and the across-scene coefficient of variation is
essentially unchanged ($19.7\%$ with, $18.8\%$ without), i.e.\ anchoring
the thresholds in meters does not by itself compress the budget spread
at benchmark scales --- its value is the protection against gross
mis-scale documented above, which these scenes never experience.

We deliberately do not report a training-time comparison from these
runs. The full-system and ablation batches were executed at different
times under different machine load (per-scene wall clock ranges
$6.4$--$32.0$\,min for one condition and $5.3$--$8.9$\,min for another
on the same scenes), so any timing difference here would measure the
queue, not the method. The controlled same-session measurement reported
in the main paper stands.

\section{Statistical detail}
\label{supp:stats}
The error distributions are heavy-tailed (two abstained scenes inherit
${\sim}70\%$ warm errors), so Table~\ref{tab:supp-ci} reports bootstrap
95\% confidence intervals (10{,}000 resamples) for every headline
statistic, and all comparative claims in the main paper rest on paired
tests. The router-vs-MASt3R-everywhere margin is $-0.5$\,pp with CI
$[-2.9,+2.0]$ ($p{=}0.36$, Wilcoxon): the router's contribution is the
60\% invocation saving at statistically equal accuracy, and we note
MASt3R is itself marginally better on our abstained scenes (9.4\%) than
on active ones (10.2\%).

An insignificant difference is not by itself evidence of equivalence, so
we additionally run a paired two-one-sided test (TOST) on the same 47
per-scene differences. We pre-specify an equivalence margin of
$\pm3$\,pp, twice the $<1.5$\,pp variation our own training re-runs
induce (Sec.~\ref{supp:locale}): a difference smaller than twice the
pipeline's own re-run noise is not one a practitioner could act on.
Under TOST both one-sided nulls are rejected ($p{=}0.028$; 90\% CI
$[-2.65,+1.58]$), so the router is statistically \emph{equivalent} to
running MASt3R everywhere within that margin, not merely
not-significantly-different. At $\pm2.5$\,pp the test does not pass, and
we report the margin at which the claim holds rather than the smallest
one we could find.

\begin{table}[htbp]
  \centering\scriptsize
  \caption{Visibility-gate ablation, ScanNet++ 47 (same trained models,
  post-hoc estimator re-run per variant; deployed protocol).
  ``active err'' is the mean over the scenes each variant chooses to act
  on, so it is not comparable across rows: removing the gate looks
  excellent on that column precisely because it only acts on the 16
  easiest scenes.}
  \label{tab:supp-vis}
  \setlength{\tabcolsep}{3.4pt}
  \begin{tabular}{lccccc}
    \toprule
    voting variant & mean & median & abstain & active & router @calls \\
    \midrule
    \textbf{one-sided (ours)} & \textbf{16.6\%} & \textbf{10.6\%} & \textbf{19} & \textbf{28} & 9.4\% @19 \\
    contribution-weighted & 17.6\% & 11.6\% & 22 & 25 & 9.8\% @22 \\
    symmetric $|z{-}D|$ & 18.7\% & 13.6\% & 27 & 20 & 9.9\% @27 \\
    no gating & 16.4\% & 12.4\% & 31 & 16 & 9.0\% @31 \\
    \bottomrule
  \end{tabular}
\end{table}

\begin{table}[htbp]
  \centering\scriptsize
  \caption{Coverage-only routing cannot replace the inlier-mass gate
  (ScanNet++ 47): routing the lowest-coverage scenes to MASt3R at the
  same invocation budget (19 calls) is far worse, and matching the gate's
  accuracy requires invoking MASt3R on nearly every scene.}
  \label{tab:supp-covroute}
  \begin{tabular}{lccc}
    \toprule
    router statistic & mean @19 calls & best mean & calls at best \\
    \midrule
    door coverage           & 15.3\% & 9.6\% & 37 \\
    ceiling coverage        & 14.5\% & 9.9\% & 47 \\
    door$+$ceiling coverage & 14.6\% & 9.6\% & 45 \\
    \textbf{inlier mass (ours)} & \textbf{9.4\%} & --- & \textbf{19} \\
    \bottomrule
  \end{tabular}
\end{table}

\begin{table}[htbp]
  \centering\scriptsize
  \caption{Bootstrap 95\% CIs, ScanNet++ 47.}
  \label{tab:supp-ci}
  \begin{tabular}{lc}
    \toprule
    statistic & value [95\% CI] \\
    \midrule
    quantile mean & 21.1 [16.9, 25.7] \\
    quantile median & 17.5 [13.9, 21.4] \\
    ours mean & 16.6 [12.2, 21.6] \\
    ours median & 10.6 [7.7, 13.9] \\
    ours active mean (28) & 9.3 [6.6, 12.4] \\
    quantile active mean (28) & 16.8 [13.3, 20.5] \\
    quantile abstained mean (19) & 27.4 [19.1, 37.0] \\
    MASt3R mean & 9.9 [7.8, 12.2] \\
    router mean & 9.4 [7.3, 11.7] \\
    \bottomrule
  \end{tabular}
\end{table}

\paragraph{Paired differences rather than marginal intervals.}
Reporting our bootstrap interval beside the baseline's invites an unpaired
reading of a paired result: the two marginal intervals overlap heavily even
where the paired difference excludes zero. The comparison the claims actually
rest on is therefore given directly. Against the quantile heuristic the paired
reduction is 4.5\,pp (95\% CI [1.9, 7.2], Wilcoxon $p{=}2.6\times10^{-3}$) over all 47
scenes, and 7.5\,pp ([3.3, 11.6], $p{=}1.9\times10^{-3}$) over the 28 scenes the
estimator acts on. The last column of the main table is that same fixed
28-scene set on every row, so it can be read vertically; each method's own
active set differs and is reported in the text instead.

\section{Instance completeness as a cue-level filter}
\label{supp:completeness}
The principal residual failure is correlated partial measurement:
truncated instances measure short, several of them agree, and together
they outvote the correct anchor, which the cue-mass gate cannot
detect, precisely because they agree. We tested whether completeness can
be estimated directly instead. Classes that rest on the floor should have
their lowest points \emph{on} the floor, so for each such instance we
take its clearance: the height of its lowest trimmed point above the
semantic floor plane, divided by its own vertical extent. The ratio is
scale-free, which matters because the scale is the unknown, and it is
independent of whether the cue agrees with the others, which is the
property the inlier mass lacks.

The signal is present in quantity: across the 47 scenes the median
clearance is $0.37$ extents and 53\% of floor-standing instance cues
exceed $0.25$, i.e.\ they are truncated from below. Filtering on it gives
a precision/coverage dial (Tab.~\ref{tab:supp-complete}; every variant is
refit on one shared set of votes, so the rows are exactly paired and the
unfiltered row reproduces the main table exactly). A mild filter is free:
dropping cues above $1.0$ improves every aggregate slightly at unchanged
coverage. An aggressive filter trades hard: at $0.1$ the error on scenes
the system still acts on falls from 9.3\% to 7.2\% and the number of
degraded scenes is more than halved, from 9 to 4, at the cost of 12
additional abstentions.

Two caveats we can state precisely. The test only detects truncation from
\emph{below}: the worst failure scene, whose doors are cut off at the
\emph{top}, is essentially unaffected ($24.6\%\!\to\!24.3\%$). And this
is an analysis of a frozen system, and adopting it would mean
re-freezing and re-validating, so we report it as a validated direction
rather than folding it into the results.

\begin{table}[htbp]
  \centering\scriptsize
  \caption{Completeness filtering, ScanNet++ 47 (paired: all rows refit on
  one shared vote draw). ``degraded'' counts active scenes ending worse
  than the warm start.}
  \label{tab:supp-complete}
  \begin{tabular}{lccccc}
    \toprule
    clearance rule & mean & median & active & abstain & degraded \\
    \midrule
    none (frozen) & 16.6\% & 10.6\% & 9.3\% & 19 & 9 \\
    drop $>1.00$ & \textbf{16.1\%} & \textbf{10.3\%} & 8.7\% & 18 & 9 \\
    drop $>0.50$ & 16.3\% & 11.1\% & 8.5\% & 21 & 7 \\
    drop $>0.25$ & 17.4\% & 12.5\% & 9.1\% & 26 & 7 \\
    drop $>0.10$ & 17.8\% & 12.9\% & \textbf{7.2\%} & 31 & \textbf{4} \\
    \bottomrule
  \end{tabular}
\end{table}

\paragraph{Adopting the filter and re-freezing.}
The paper does not adopt this filter, because its rule was chosen while
looking at all 47 scenes. To say what adopting it would actually buy, we ran
the selection honestly: the rule (scope, drop-or-downweight, $\tau$) is chosen
on the 19 development scenes alone and then applied unchanged to the 28
held-out scenes. Two selection criteria were run rather than the flattering
one --- fewest degraded scenes, and lowest deployed mean --- since they are
different questions and only the first is what the filter was designed for.

Under the `fewest degraded' criterion the selected rule (all/drop/tau=0.1) reproduces on
held-out data exactly the trade the paper reports: degraded scenes fall from
5 to 2, but abstentions rise from 15 to 20 and the deployed mean does
not move (20.5\% to 20.2\%, paired -0.28\,pp, 95\% CI [-1.97, +1.51], $p{=}0.75$).
Under the `lowest deployed mean' criterion the selected rule is a near
no-op on held-out data (20.5\% to 20.5\%, $p{=}0.86$). Over all 47 scenes the
first rule turns 9 degraded scenes into 4 at the cost of 12 additional
abstentions while the deployed mean moves from 16.6\% to 17.8\%.

The conclusion we draw is stronger than the discipline argument we could have
made: the completeness signal converts degradations into refusals rather than
into accuracy, so adopting it would not have changed the headline.

\begin{figure}[htbp]
  \centering
  \includegraphics[width=0.86\linewidth]{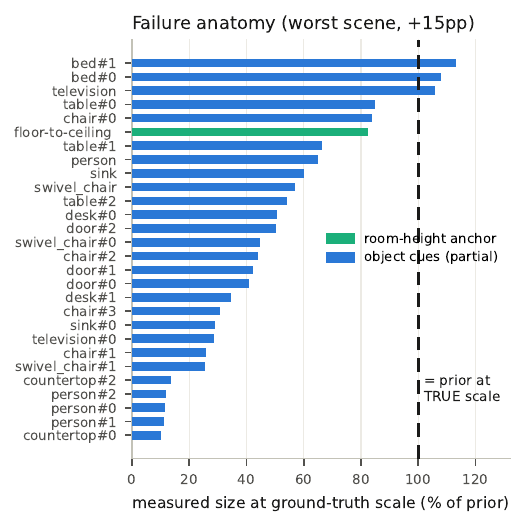}
  \caption{\textbf{Anatomy of the worst failure.} Cue sizes at the
  ground-truth scale, as a fraction of their priors: nearly all object cues
  are partial (doors at 41--50\%), so the robust fusion inflates the scale
  past the one correct anchor (room height, 82\%).}
  \label{fig:supp-failure}
\end{figure}

\section{Cost of a wrong vertical axis}
\label{supp:tilt}
Every extent is measured along one axis, so a tilted axis attenuates all
of them by the cosine. To measure what that costs we perturbed the
selected axis by a controlled angle and re-measured everything else
identically, reusing each scene's votes across angles so the rows are
exactly paired; the zero-tilt row reproduces the main table to
$0.00$\,pp.

Small tilts are free and large ones are caught rather than silently
absorbed (Tab.~\ref{tab:supp-tilt}). Up to $5^\circ$ the aggregates move
within their own noise. At $10^\circ$ the mean rises 1.4\,pp, and beyond
that the degradation is real: $+3.5$\,pp at $15^\circ$ and
$+9.2$\,pp at $25^\circ$, the latter close to the $9.4\%$ attenuation
$1-\cos 25^\circ$ predicts. Crucially, abstentions rise with it, from 19
to 29: a wrong axis shrinks every extent at once, the cues stop agreeing
with the room-height prior, and the gate fires. Axis error therefore
surfaces mainly as additional refusals, which is the behavior the gate
exists to provide.

\begin{table}[htbp]
  \centering\scriptsize
  \caption{Vertical-axis tilt vs.\ scale error (ScanNet++ 47, paired:
  votes computed once per scene and reused across angles).}
  \label{tab:supp-tilt}
  \begin{tabular}{lccc}
    \toprule
    tilt & mean & median & abstain \\
    \midrule
    $0^\circ$ (selected) & 16.6\% & 10.6\% & 19 \\
    $2^\circ$ & 17.4\% & 11.5\% & 19 \\
    $5^\circ$ & 16.8\% & 10.2\% & 18 \\
    $10^\circ$ & 18.0\% & 12.4\% & 22 \\
    $15^\circ$ & 20.1\% & 13.9\% & 29 \\
    $25^\circ$ & 25.8\% & 16.8\% & 29 \\
    \bottomrule
  \end{tabular}
\end{table}

\section{Modeling the dependency between cues}
\label{supp:latent}
Our fusion treats cues as independent, which is wrong in exactly the way
that hurts: partial reconstruction biases them together. We tested the
simplest dependency model that captures this: a single latent
per-scene coverage factor $\kappa\in(0,1]$ shared by all instance-extent
cues, so an instance is expected to measure $\kappa\mu_c$ rather than
$\mu_c$, fitted jointly with the scale on a 2D grid under a prior pulling
$\kappa$ toward 1. Room-height cues are excluded from $\kappa$: a gap
between two planes is not an object extent and is not truncated the same
way, and that asymmetry is what keeps the model identifiable at all.

What it buys is coverage: with a moderate prior
($\sigma_\kappa{=}0.2$) the fitted factor is 0.94 in the median (0.74 at
the extreme), so the model does detect systematic under-measurement,
and abstentions fall from 19 to 15 while the mean improves from 16.6\% to
15.7\%; but degraded scenes do not fall (9 to 10). Loosening the prior
further ($\sigma_\kappa{=}0.4$) continues that trade, reaching 13
abstentions at 16.2\%.

The reason is structural, and it sharpens the contrast with
Sec.~\ref{supp:completeness}: a scene-global $\kappa$ is nearly
degenerate with the scale itself (shrinking every object prior by
$\kappa$ and enlarging $s$ by $1/\kappa$ is almost the same fit), so
only the room-height cue breaks the tie, and in exactly the scenes that
fail, that cue is the one being outvoted. Because the bias is fitted
from the same cues that carry it, the fit cannot recover the truth; the
completeness signal works better because it brings \emph{external}
geometric evidence about whether an instance is whole.

\section{How much is the result bounded by the teacher?}
\label{supp:teacher}
Every label in the paper comes from one frozen segmenter (Mask2Former
R50, 47.2 mIoU on ADE20K). To separate ``limited by its teacher'' from
``limited by partial reconstruction'' we re-ran the post-hoc estimator on
the same trained models, with the same seeds, priors and gate, swapping
in Mask2Former Swin-L (57.3 mIoU). Nothing else changed.

The stronger teacher helps, and it helps most where the method is
weakest (Tab.~\ref{tab:supp-teacher}). On the 24 scenes both teachers
act on, the like-for-like comparison, the mean error falls from
9.7\% to 6.3\%. It repairs the failure case dissected in the main
paper, whose doors were measuring
41--50\% of their prior: that scene goes from $24.6\%$ (a $15$\,pp
degradation) to $8.6\%$, and degraded scenes overall fall from 9 to 6
while abstentions fall from 19 to 17. Interestingly the cue \emph{count}
barely moves (41 to 39 per scene on average); the gain comes from the
completeness of each labeled instance.

This qualifies the failure analysis in the main paper: we attributed the
worst errors to partial reconstruction, but they are partly partial
\emph{labeling}, and a better segmenter recovers much of the gap. The
all-scene mean improves less dramatically ($16.6\%\!\to\!15.7\%$,
$p{=}0.16$) because it is dominated by abstained scenes, which inherit
the warm start under either teacher. We report this as a sensitivity
study rather than adopting Swin-L: the configuration was frozen before
public evaluation, and swapping a component would forfeit that.

\begin{table}[htbp]
  \centering\scriptsize
  \caption{Teacher sensitivity, ScanNet++ 47 (same models, same seeds,
  same gate; only the segmenter differs).}
  \label{tab:supp-teacher}
  \begin{tabular}{lcc}
    \toprule
    & R50 (47.2 mIoU) & Swin-L (57.3 mIoU) \\
    \midrule
    mean & 16.6\% & \textbf{15.7\%} \\
    median & 10.6\% & \textbf{8.7\%} \\
    both-act subset (24) & 9.7\% & \textbf{6.3\%} \\
    active mean (28 / 30 scenes) & \textbf{9.3\%} & 10.3\% \\
    worst failure scene & 24.6\% & \textbf{8.6\%} \\
    scenes acted on & 28 & 30 \\
    degraded & 9 & \textbf{6} \\
    \bottomrule
  \end{tabular}
\end{table}

\section{Uncertainty calibration beyond the gate}
\label{supp:uncertainty}
The abstention gate is a binary decision; a natural question is whether a
continuous confidence would serve better. We bootstrap each scene's cue
set (200 resamples, refitting the robust MLE each time) and take the
relative width of the resulting 95\% interval on the residual scale as a
per-scene confidence.

The interval behaves sensibly across regimes (its median width is
0.403 on abstained scenes versus 0.137 on active ones) and correlates
with the true error over all 47 scenes ($\rho{=}{+}0.40$, $p{=}0.005$).
It is nonetheless a \emph{worse} signal than the inlier mass we already
use ($\rho{=}{-}0.57$, $p{<}0.001$), and routing by interval width at the
gate's own budget of 19 MASt3R calls gives 11.0\% mean error against the
gate's 9.4\% (the two selections agree on only 11 of 19 scenes).

The more important negative is that \emph{within} the 28 active scenes
neither statistic predicts the error (interval width $\rho{=}{+}0.18$,
$p{=}0.36$; inlier mass $\rho{=}{-}0.11$, $p{=}0.59$). This is the same
two-level structure reported for coverage in the main paper, and it is
consistent with the failure anatomy: a scene whose partial cues agree
with one another is confident \emph{and} wrong, so no statistic computed
from cue agreement, including a bootstrap interval, can separate it.
Calibrated per-scene confidence therefore needs evidence the fusion does
not currently have: a signal for whether an instance is \emph{complete},
independent of whether its measurement agrees with the others.

\section{Other benchmarks: ETH3D per scene}
\label{supp:eth3d}
ETH3D (sim3 protocol, frozen configuration) probes the regime boundary
(Table~\ref{tab:supp-eth3d}): one indoor office is corrected actively;
the outdoor courtyard and the two industrial rooms abstain, the
outdoor scene with inlier mass 0.5, far below the gate.
One of the four falls inside the operating regime the main paper defines
--- real photographs of a room whose architecture is visible --- and that
is \texttt{office}; the courtyard is outdoor and the other two are
industrial halls, and the gate refuses all three. On that one in-regime
scene MoGe-2 is the more accurate estimator (1.9\% against our 9.1\%). We
report the whole table rather than the single row (\texttt{pipes}) that
favors us. One in-regime scene is not a benchmark verdict and we do not
treat it as one, but it is the only uncontaminated public test of this
kind available to us and it did not go our way.

\begin{table}[htbp]
  \centering\scriptsize
  \caption{ETH3D per-scene scale error (\%). MoGe-2 runs the same
  alignment recipe; ETH3D is outside its training mixture. \texttt{delivery\_area} is the outdoor scene.}
  \label{tab:supp-eth3d}
  \setlength{\tabcolsep}{3pt}
  \begin{tabular}{lccccc}
    \toprule
    scene & quantile & ours & MoGe-2 & mass & decision \\
    \midrule
    delivery\_area & 51.3 & 51.3 & \textbf{3.7} & 0.5 & abstain \\
    kicker & 14.4 & 14.4 & \textbf{0.3} & 1.4 & abstain \\
    office & 16.5 & 9.1 & \textbf{1.9} & 1.9 & active \\
    pipes & 9.6 & \textbf{9.6} & 16.5 & 1.0 & abstain \\
    \bottomrule
  \end{tabular}
\end{table}

\section{A benchmark inside a model's training set}
\label{supp:indomain}
Run as a scale estimator on ScanNet++, MoGe-2 reaches 2.3\% mean and 1.9\%
median error, ahead of every other method in this paper by a wide margin, with
45 of 47 scenes under 5\%. That result is real but it is not a like-for-like
comparison, and reading it as one would be a mistake: ScanNet++ appears in MoGe-2's own training
mixture (its Tab.~A.1 lists it at 176K frames and 4.6\% sampling weight).
Those scenes are in-domain for MoGe-2 and held out for us.

The comparison that matters is therefore on data outside that mixture, and we
run MoGe-2 unchanged on three such sets rather than argue from one.
Tab.~\ref{tab:supp-indomain} gives the five development captures against all
three metric-depth networks under the identical alignment protocol and shared
measurement operator; Tab.~\ref{tab:supp-ood} summarizes the picture across
every set we can evaluate.

The result is not a clean win for either side and we report it as it came out.
On ETH3D --- DSLR imagery, laser reference, and not in MoGe-2's training list
--- MoGe-2 is clearly better than us (5.6\% against our 21.1\%, and we act on
only one of its four scenes). On ordinary captures nobody curated it is not:
34.8\% against 7.7\% on the development captures, and 28.4\% against 12.3\% on
the 13 in-the-wild captures of Sec.~4.7, where it wins 3 of 9 scenes, lands
level with the quantile heuristic it would replace (25.2\%), and produces
outright failures a practitioner would notice (a corridor at $-69.7\%$, an
apartment at $+51.5\%$). Its error grows by an order of magnitude moving from
a benchmark inside its training corpus to consumer captures outside it.

Two conclusions. The three sets form an exposure ladder --- inside the
training mixture (2.3\%), a benchmark the model was evaluated against
(5.6\%), and captures it has never seen (28--35\%) --- but exposure and
capture grade co-vary along it (ScanNet++ is DSLR and phone, ETH3D is DSLR,
ours are phones) and we cannot separate the two. Under either reading a
published metric-scale number is a best case rather than a deployment
estimate. And on
benchmark-grade imagery a strong feed-forward model is the better choice than
what we propose; our estimator's case rests on the ordinary captures, where it
is the more reliable of the two and, unlike MoGe-2, says when it is not
confident.

\begin{table}[htbp]
  \centering\scriptsize
  \caption{MoGe-2 as a scale estimator, in and out of its training
  distribution. Mean $|$error$|$; ``ours'' is the deployed frozen system on the
  same scenes. Only the first row is inside MoGe-2's training mixture.
  $^{*}$Nine of the 13 captures of the main paper's Tab.~2: the
  four excluded ones deliver no floor$\to$ceiling height for any method
  (the robotics lab and one phone capture never frame a ceiling, the
  aerial survey yields no cue at all, and the sports field no
  floor$\to$ceiling cue), so no column can be scored on them.
  $^{\dagger}$ETH3D is not training data for MoGe-2 but is one of the
  benchmarks it reports against, so it sits between the other rows on the
  exposure ladder rather than at the unseen end.}
  \label{tab:supp-ood}
  \setlength{\tabcolsep}{4pt}
  \begin{tabular}{llcc}
    \toprule
    set & in MoGe-2 training? & MoGe-2 & ours \\
    \midrule
    ScanNet++ (47) & \textbf{yes} (176K fr., 4.6\%) & \textbf{2.3\%} & 16.6\% \\
    ETH3D (4)      & no (its eval set)$^{\dagger}$  & \textbf{5.6\%} & 21.1\% \\
    development captures (5) & no & 34.8\% & \textbf{7.7\%} \\
    in-the-wild captures (9)$^{*}$ & no & 28.4\% & \textbf{12.3\%} \\
    \bottomrule
  \end{tabular}
\end{table}

\begin{table}[htbp]
  \centering\scriptsize
  \caption{Metric-depth alignment on the development captures, which no
  released model has trained on. Signed delivered ceiling-height error
  against the tape measure; identical protocol and measurement operator for
  every column.}
  \label{tab:supp-indomain}
  \setlength{\tabcolsep}{3.2pt}
  \begin{tabular}{lccccc}
    \toprule
    scene & quantile & DA-3 & Depth Pro & MoGe-2 & ours \\
    \midrule
    \texttt{room1} & -39.1\% & -62.4\% & -78.1\% & -50.8\% & -15.1\% \\
    \texttt{living\_1} & -15.7\% & -68.9\% & -65.5\% & -25.8\% & +8.5\% \\
    \texttt{living\_2} & -15.7\% & -60.8\% & -65.5\% & -25.8\% & +6.7\% \\
    \texttt{room2} & -46.3\% & -74.6\% & -77.0\% & -42.1\% & -5.5\% \\
    \texttt{studio1} & -20.5\% & -64.6\% & -68.5\% & -29.4\% & +3.0\% \\
    \midrule
    mean $|$err$|$ & 27.4\% & 66.3\% & 70.9\% & 34.8\% & \textbf{7.7\%} \\
    \bottomrule
  \end{tabular}
\end{table}

\section{Cost accounting}
\label{supp:cost}
Table~\ref{tab:supp-cost} breaks down the post-hoc estimator's wall
clock on a representative 47-scene model (337k Gaussians, RTX~4080),
with \emph{nothing} amortized: teacher inference is run cold on every
unique view drawn by the three seed ensembles, and the visibility test's
depth renders are included. As in the main paper, 3DGS training is
excluded as a sunk cost of the assumed deployment; a user who needs only
the scale is better served by MASt3R end to end. During training, the same teacher outputs
are shared with the in-loop voting through a cache, which is why the
scale machinery adds no measurable training overhead. The MASt3R
comparison point is its $k{=}12$ sparse global alignment
(${\sim}4$\,min on the same GPU); doubling its frame budget quadruples
the pairwise cost for a ${\sim}1$\,pp gain.

\begin{table}[htbp]
  \centering\scriptsize
  \caption{Post-hoc estimator cost breakdown (single scene, cold).}
  \label{tab:supp-cost}
  \begin{tabular}{lr}
    \toprule
    component & wall clock \\
    \midrule
    model + camera loading & 6.2\,s \\
    teacher 2D inference (127 unique views) & 3.9\,s (31\,ms/view) \\
    rendered-depth visibility ($3{\times}48$ renders) & 0.2\,s \\
    vote projection + instance measurement (3 seeds) & 0.6\,s \\
    robust MLE (2001-point grid) & $<$0.01\,s \\
    \midrule
    total & ${\sim}11$\,s \\
    MASt3R $k{=}12$ sparse global alignment & ${\sim}240$\,s \\
    \bottomrule
  \end{tabular}
\end{table}

\section{Three directions we scoped but did not adopt}
\label{supp:scoped}
\paragraph{A data-driven warm start.} The nominal room height is the one
sensitive input, so replacing it with a per-scene estimate is tempting.
We tried the obvious version (estimate the room height from the object
cues alone, shrink the prior center toward it by the object cue mass, and
refit) and it does not remove the sensitivity: across assumed heights
of 2.1--3.0\,m the spread of the mean error widens slightly (12.4 to
13.4\,pp) while medians and abstention counts improve. The failure has
the same cause as Sec.~\ref{supp:latent}: object cues carry the very
truncation bias that the room-height prior is there to stabilize, so
deriving the prior from them removes the independent evidence.

\paragraph{Finer visibility reasoning.} Our gate is a one-sided depth
test, deliberately cheap. The mask-level occlusion analysis of LabelGS
and the per-ray marginal-contribution weighting of VALA (both cited in
the main paper) reason about
visibility far more precisely, and Sec.~\ref{supp:teacher} shows why that
is the promising direction: what limits us is not how many instances we
find but how complete each labeled instance is. We have not combined
them, and we would expect the gain to overlap with the teacher upgrade
rather than add to it.

\paragraph{Sub-scene routing.} Routing is currently per scene. Routing
per region, invoking the heavy model only where local cue mass is low,
is attractive, but our drift measurement bounds the benefit: over a
703-frame corridor the per-region floor-to-ceiling gap varies by only
3\% (CV), so within a room-to-corridor extent there is little spatial
heterogeneity for sub-scene routing to exploit. Multi-floor or
multi-building captures, where a single global scale is itself the wrong
model, are the case that would justify it.

\section{What the error means in practice}
\label{supp:apps}
\begin{figure}[htbp]
  \centering
  \includegraphics[width=0.92\linewidth]{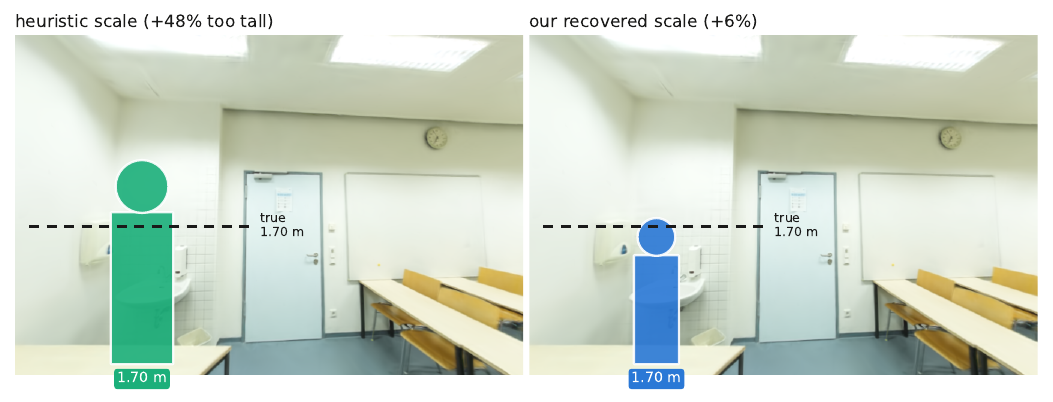}
  \caption{Metric scale is what AR sees. A $1.70$\,m mannequin
  composited into the reconstruction at each hypothesis' scale: the
  heuristic renders a $+48\%$ giant; ours matches the true height
  (dashed).}
  \label{fig:supp-mannequin}
\end{figure}

\begin{figure}[htbp]
  \centering
  \includegraphics[width=0.85\linewidth]{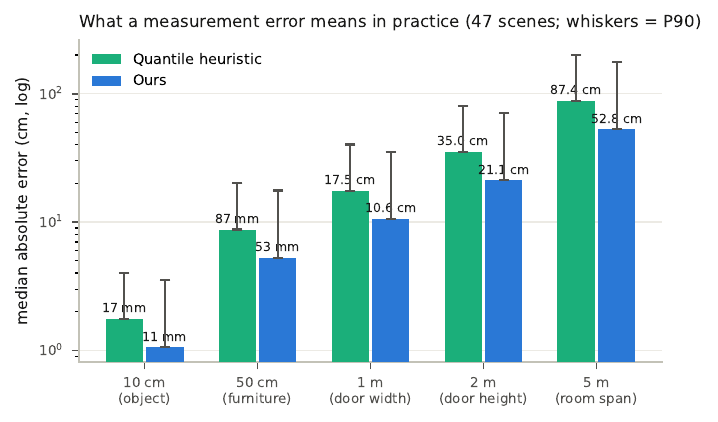}
  \caption{Measurement error by length scale, derived from the 47
  per-scene scale errors.}
  \label{fig:supp-bands}
\end{figure}

Two figures translate a percentage scale error into the terms an
application actually cares about. Fig.~\ref{fig:supp-mannequin}
composites a $1.70$\,m mannequin into a reconstruction at each scale
hypothesis: under the quantile heuristic the figure renders as a
$+48\%$ giant, while under ours it matches the true height. This is the
error an AR placement inherits directly.
Fig.~\ref{fig:supp-bands} converts the 47 per-scene errors into absolute
terms by length: because a global similarity leaves relative error
length-independent, a door measured in the model errs by $21$\,cm at the
median under our scale against $35$\,cm under the heuristic, and a
$10$\,cm object by $11$\,mm against $17$\,mm.

\paragraph{Converting scale error into decisions.}
Table~\ref{tab:supp-clearance} plays out a concrete task on the saved
per-scene delivered scales: for each of the 47 scenes and five
object/aperture pairs, the aperture is read off each method's delivered
model and the fit decision is compared with the ground-truth-scale
decision. The router flips 24.7\% of clearance judgements against
37.0\% for the quantile heuristic, 28.5\% for MASt3R everywhere, and
43.0\% for Depth Anything~3: it removes a third of the errors the
heuristic makes and still edges out the heavy fallback it invokes. None
of the methods is yet accurate enough for a clearance decision to be
taken unsupervised --- the 2.5\%-margin pair (78\,cm against an 80\,cm
doorway) defeats every method, which is consistent with
Fig.~\ref{fig:supp-bands}.

\begin{table}[htbp]
  \centering\scriptsize
  \caption{Decision flips over five object/aperture pairs per scene (an
  80\,cm doorway against objects of 60/75/78/85/90\,cm; true verdicts
  pass/pass/pass/fail/fail), ScanNet++ 47. Signed delivered scales;
  abstentions inherit the warm estimate (ours) or are routed to MASt3R
  (router). $^{\ddagger}$ScanNet++ is in MoGe-2's training corpus
  (Sec.~\ref{supp:indomain}); its row is in-domain and not comparable.}
  \label{tab:supp-clearance}
  \setlength{\tabcolsep}{4pt}
  \begin{tabular}{lcc}
    \toprule
    method & flipped decisions & scenes with any flip \\
    \midrule
    quantile heuristic & 87/235 (37.0\%) & 44/47 \\
    DA-3 alignment & 101/235 (43.0\%) & 43/47 \\
    MoGe-2 alignment$^{\ddagger}$ & 11/235 (4.7\%) & 9/47 \\
    MASt3R (metric) & 67/235 (28.5\%) & 38/47 \\
    Ours (frozen) & 69/235 (29.4\%) & 40/47 \\
    Ours\,+\,MASt3R router & \textbf{58/235 (24.7\%)} & \textbf{36/47} \\
    \bottomrule
  \end{tabular}
\end{table}

\section{Custom captures: protocol and per-capture detail}
\label{supp:custom}
This section backs Sec.~4.7 of the main paper. The 13 captures were
recorded with two Android handsets, two iPhones and a consumer drone,
and reconstructed with the same COLMAP settings as every other
real-SfM condition in this paper. Training and estimation used the
frozen configuration verbatim: nominal height $2.5$\,m, $\lambda{=}1$,
$\sigma_w{=}0.35$, $M_{\min}{=}1.5$, the R50 teacher, three vote seeds,
and the depth regularizer at $0.05$. No per-capture tuning of any kind
was performed, and the captures were evaluated once.

\paragraph{Why Tab.~2 has no MASt3R column.}
We ran MASt3R's metric checkpoint on all 13 captures under the paper's
own protocol (evenly sampled frames, sparse global alignment, the
shared measurement operator). At the frame budgets we could afford
($k{=}12$ and $k{=}24$) its sparse global alignment meets the 20\,cm
registration gate of protocol P2 on only one of the 13 captures
(RMSE 0.5--4.1\,m on the other twelve at $k{=}24$, against
${\sim}0.07$\,m on ScanNet++ at $k{=}12$), so no scale of
ours-versus-its kind is measurable on them under our own protocol, and
tabulating the resulting numbers (mean $|$error$|$ 66--71\%, every
capture under-scaled) would score a reconstruction that never
registered rather than MASt3R's scale. On the one capture that does
pass the gate (a single-dwelling session, RMSE 6.6\,cm at $k{=}24$)
its error is $+13.1\%$ --- consistent with its ScanNet++ behavior ---
against our $+2.6\%$. The heavy fallback the router assumes therefore
has a scene-extent precondition that room-scale benchmarks do not
expose: a multi-room walkthrough needs a frame budget that grows with
scene extent, and the pairwise cost grows quadratically with it.

\paragraph{On the ground truth.} The capture owner reports a nominal
ceiling height per venue class rather than measuring each room:
$2.3$\,m for the dwellings and show flat, $3.0$\,m for the lab and
office floor. This is the same kind of approximate reference as the
development captures of the main paper, and it is weaker than the laser
ground truth used for ScanNet++, so we report delivered height rather
than scale error and we do not aggregate these numbers with the public
benchmarks. Real ceilings vary within a class, which sets a floor on the
error any method can show here.

\begin{figure}[htbp]
  \centering
  \includegraphics[width=\linewidth]{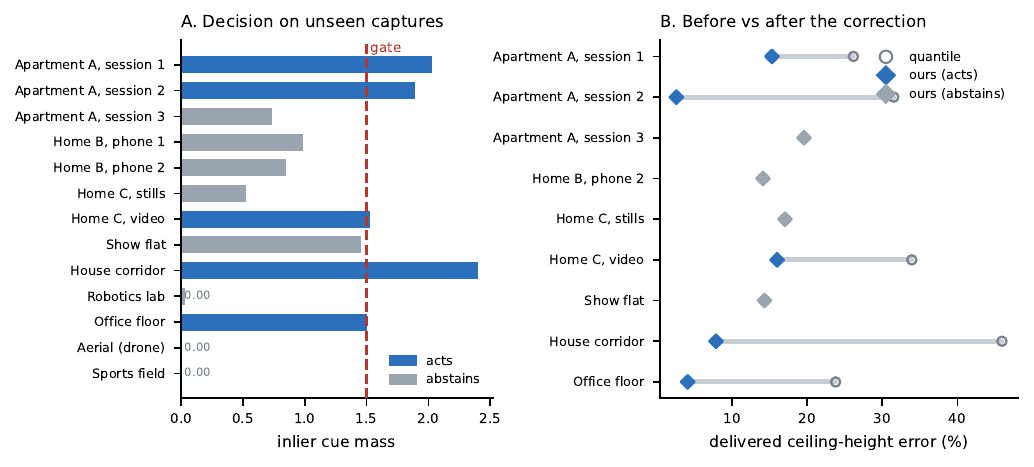}
  \caption{\textbf{Frozen system on our own captures.} Left: inlier cue
  mass against the frozen gate. Right: delivered ceiling-height error
  before (circles) and after (diamonds) the semantic correction. With
  only a nominal per-venue reference these are transfer evidence, not an
  accuracy result.}
  \label{fig:supp-custom}
\end{figure}

\paragraph{Repeat captures.} Three of the venues were recorded more than
once, which gives a ground-truth-free consistency check. Apartment A was
captured in three separate sessions: the two richer sessions clear the
gate and deliver $2.65$ and $2.36$\,m, while the sparsest (167 images
against 376) abstains at mass $0.73$ and inherits $2.75$\,m. Home C was
captured as stills and as video; the video clears the gate and is
corrected to $2.67$\,m while the stills abstain. Coverage, not venue,
decides: within one dwelling the decision tracks how much of the
architecture the capture actually shows.

\paragraph{Why the refusals happen.} The eight abstentions split into
three mechanisms. The sports field and the aerial survey end at mass
$0.00$: the drone capture yields no measurement to fuse at all (the
degenerate limit of the same test), and the six cues the sports field
does yield are all outliers to the warm prior, so none contributes
mass. The
robotics lab never frames its ceiling, leaving nine cues and mass
$0.03$. The remaining five are cue \emph{disagreement} rather than cue
absence: 21--39 cues survive, but partial furniture and mislabeled
instances scatter them, exactly the mechanism dissected for handheld
video in Sec.~4.6 of the main paper. In one apartment capture a coat
stand is labeled \texttt{person} and measures $3.7$\,m against a
$1.70$\,m prior; the gate fires instead of absorbing it. That anchor is
the weakest in the catalog and a fair thing to question. Post hoc, on
the frozen cue sets, dropping \texttt{person} entirely moves ScanNet++
from 16.6\% to 16.8\% mean with the median unchanged and one fewer
abstention: it is close to inert on that benchmark, it costs nothing to
keep, and removing it now would break the freeze the rest of the paper
depends on. We keep it and report the failure it can produce, since it
shares its $1.70$\,m mean with \texttt{refrigerator} and the two can
reinforce each other under misclassification.

\paragraph{Does the weak reference decide the comparison?}
The captures of Sec.~4.7 are scored against a nominal per-venue ceiling rather
than a tape measure, and real ceilings vary within a class. The honest question
is not what the exact error is but whether the comparison survives the
reference being wrong, so we sweep the assumed true dwelling height across the
range ordinary dwellings span and rescore both methods against it. Nothing is
re-estimated: the delivered heights are fixed outputs and only the reference
moves.

Table~\ref{tab:supp-gtsens} gives the result. Our mean error is below MoGe-2's
at every point in the range, and the two never come closer than 6.7\,pp. The
$2.3$\,m we report is nearer the unfavorable end for us than the favorable
one: at $2.5$--$2.6$\,m we would report $8$--$10\%$ rather than $13\%$. Per
capture, the height at which the two methods are equally close lies outside
the plausible range in seven of the nine scoreable captures --- the eight
dwellings of this table plus the office floor, which carries the $3.0$\,m
nominal --- so the pairwise verdict is not a function of the reference
either. This does not turn a nominal reference into
a measurement --- a tape measure would still be worth having --- but it does
mean the in-the-wild conclusion is not an artefact of it.

\begin{table}[htbp]
  \centering\scriptsize
  \caption{Delivered ceiling-height error on the dwelling captures as a
  function of the assumed true height. Both columns are rescored from fixed
  delivered heights; nothing is re-estimated.}
  \label{tab:supp-gtsens}
  \begin{tabular}{lccc}
    \toprule
    assumed truth (m) & ours & MoGe-2 & ours closer \\
    \midrule
    2.1 & 19.6\% & 26.3\% & 4/8 \\
    2.2 & 16.2\% & 26.2\% & 4/8 \\
    2.3 & 13.3\% & 27.1\% & 5/8 \\
    2.4 & 11.1\% & 28.0\% & 6/8 \\
    2.5 & 9.7\% & 29.9\% & 7/8 \\
    2.6 & 8.4\% & 31.6\% & 7/8 \\
    2.7 & 9.2\% & 33.2\% & 7/8 \\
    \bottomrule
  \end{tabular}
\end{table}

\paragraph{Limitations of this study.} Five acting captures is a small
sample, the ground truth is class-level, and the two largest wins are
both cases where the warm start fails badly, which is where any
correction has the most room to help. We report the study as evidence
that the frozen configuration transfers, not as a second benchmark.

\section{Reproducibility}
Every reported number derives from a manifest mapping (condition, scene)
$\to$ trained model, shipped with the code, together with the evaluation
scripts and the seeded benchmark converters (the applied sim(3) transforms
are reproducible from scene-name hashes).

\end{document}